\documentclass{article}
\usepackage{jfrExamplee}
\usepackage{graphicx}
\usepackage{apalike}
\usepackage{amsfonts,amssymb,amsmath,amsthm}
\usepackage{setspace}
\usepackage{color}
\usepackage{subfigure}
\usepackage{multirow}
\usepackage{url}
\newcommand{\Hbf}{\mathbf{H}}\newcommand{\Ibf}{\mathbf{I}}\newcommand{\Jbf}{\mathbf{J}}\newcommand{\Pbf}{\mathbf{P}}\newcommand{\Rbf}{\mathbf{R}}
\newcommand{\bbf}{\mathbf{b}}\newcommand{\dbf}{\mathbf{d}}\newcommand{\ebf}{\mathbf{e}}\newcommand{\lbf}{\mathbf{l}}\newcommand{\nbf}{\mathbf{n}}\newcommand{\pbf}{\mathbf{p}}\newcommand{\qbf}{\mathbf{q}}\newcommand{\rbf}{\mathbf{r}}\newcommand{\vbf}{\mathbf{v}}\newcommand{\xbf}{\mathbf{x}}\newcommand{\zbf}{\mathbf{z}}

\newcommand{\eg}{\textit{e.g.}, }
\newcommand{\ie}{\textit{i.e.}, }

\newcommand{\etc}{\textit{etc.} }

\newtheorem{remark}{Remark}

\title{SPINS: A Structure Priors aided Inertial Navigation System}

\author{
	Yang ~Lyu \thanks{Yang
		Lyu, Thien-Minh Nguyen, Muqing Cao, Shenghai Yuan, Thien Hoang Nguyen and Lihua Xie (corresponding author) are
		with School of Electrical and Electronic Engineering, Nanyang Technological
		University, 50 Nanyang Avenue, Singapore 639798. 
	} \\
	\And
	Thien-Minh Nguyen {$^*$}\\
	\And
	Liu Liu\thanks{Liu Liu is with College of Engineering and Computer Science, Australian National University, Canberra ACT 0200, Australia.}
	\And
	Muqing Cao{$^*$}
	\And
	Shenghai Yuan{$^*$}
	\And
	Thien Hoang Nguyen {$^*$}
	\And
	Lihua Xie{$^*$}
}

\begin{document}
	
	\maketitle
	
	\begin{abstract}
		We propose a navigation system combining sensor-aided inertial navigation and prior-map-based localization to improve the stability and accuracy of robot localization in structure-rich environments. 
		Specifically, we adopt point, line and plane features in the navigation system to enhance the feature richness in low texture environments and improve the localization reliability. 
		We additionally integrate structure prior information of the environments to constrain the localization drifts and improve the accuracy. 
		The prior information is called \textit{structure priors} and parameterized as low-dimensional relative distances/angles between different geometric primitives. 
		The localization is formulated as a graph-based optimization problem that contains sliding-window-based variables and factors, including IMU, heterogeneous features, and structure priors. A limited number of structure priors are selected based on the information gain to alleviate the computation burden. Finally, the proposed framework is extensively tested on synthetic data, public datasets, and, more importantly, on the real UAV flight data obtained from both indoor and outdoor inspection tasks. The results show that the proposed scheme can effectively improve the accuracy and robustness of localization for autonomous robots in civilian applications.
	\end{abstract}
	
	\section{Introduction}
	
	Autonomous vehicles, especially Unmanned Aerial Vehicles (UAVs), have attracted tremendous research interests in recent years due to their potential in improving efficiency
	and safety in many military and civilian applications \cite{cai2014survey}. 
	{\color{blue} One fundamental prerequisite for an autonomous vehicle to successfully execute missions is to accurately and reliably estimate its 6-DOF pose \cite{Zhang2015Visual}, which requires delicate algorithm development and systematic integration. A detailed review of various localization systems is provided in \cite{yuan2021survey}.}
	Due to physical and electromagnetic interferences, GPS systems may not provide persistent and reliable localization information in many complex environments, named as GPS-denied environments, making it a challenging task for a vehicle to carry out missions autonomously.
	
	There are mainly two approaches to handle the localization in a GPS-denied environment. The first approach is the simultaneous localization and mapping (SLAM) framework \cite{durrant2006simultaneous}, which is to incrementally track the local pose by estimating the relative transformation between two observation frames. The main drawback of a SLAM-based method is that the localization result drifts as time goes on due to the accumulated relative pose estimation errors \cite{strasdat2010scale}. {\color{blue} One may reason that causes the relative pose estimation error is that the salient point features, on which many localization methods are dependent, are insufficient in many challenging civilian environments.}
	{\color{blue}Line  and  plane  features  are  considered  as  promising supplements to point features for robot localization, especially in many civilian environments.} Line  and  plane  features  are  considered  as  promising supplements to point features for robot localization. First, lines and planes are more structurally salient and therefore can be endowed with more prior information. Secondly, they are more stable than points with respect to lighting/texture changes. Finally, lines and planes are more ubiquitous in a structure rich environment, such as most infrastructures in urban cities.  Thus, it is a good practice to additionally integrate line and plane features in the  localization framework for many civil applications.

	Another alternative is to localize the vehicle according to a prior map by matching the local observations to the map directly \cite{sattler2011fast}. Although this method has no drift issue, it is not easy to apply this method solely to achieve reliable localization in many challenging environments.  {\color{blue}First, the prior-map-based localization requires a high-fidelity map that may not be available in many scenarios. Second, even if there is a prior map, the association between local observations and map information is usually sophisticated and time-consuming  due to large differences in observation view angle, data resolution, and even data format \cite{sattler2018benchmarking,muhlfellner2016summary}.}

%	 In addition to the above challenges, the prior information may be in large amounts and very redundant. 
%	It is important to determine what kind of prior information is to be integrated into the local SLAM, which already poses a heavy burden to the onboard computation resources of the autonomous vehicle. 

	Reflecting on the pros and cons of the two distinct approaches, it is a good practice to combine them, namely, to lend some prior information of the map to aid the SLAM based navigation. A loosely coupled framework is proposed in \cite{platinsky2020collaborative}, which incorporates global pose factors in the local SLAM optimization. The global pose is obtained by matching images with a global map. Article \cite{middelberg2014scalable} uses a locally pre-stored 3D points cloud map to fix the drift of a local SLAM system in a tightly coupled framework. Similar works are also presented in \cite{Zuo2020Multi,mur2017visual} based on different sources of map. {\color{blue}Although the combination methods above demonstrates improved localization performances, their frequent and indiscriminate map prior information integration may severely slow down the localization process. Therefore, it is important to determine what information to integrate into the SLAM process to make a balance between localization performance and efficiency.}

	{\color{blue}Inspired by the discussions above, this paper proposes a Structure Priors aided Inertial Navigation System (SPINS) that combines SLAM-based and prior map-based localization method by lending feature level structure prior information to restrain the drift of the SLAM-based methods. To deal with the challenges such as 
	feature deficiency, prior information association, and to achieve an efficient combination of the two methods in civilian environments with rich structure information, we make the following contributions:}

%	The SPINS is a tightly-coupled framework that combines SLAM and prior map based localization by lending feature level structure prior information to restrain the drift of the SLAM based methods. 
	
%	To solve the first problem, we broaden the sources of prior information from maps to more generalized prior knowledge, ranging from the elaborated CAD blueprints to some coarse, shared knowledge, such as parallel or orthogonal lines/planes. To build associations between the prior information and the observed point, line, and plane features, rather than carrying out feature matching based on high dimensional descriptors\cite{piasco2018survey}, we parameterize the prior information as pairwise distances and angles between different geometric primitives, so that the association and integration to SLAM can be straightforward based on thresholds.
%	To reduce the computation burden of incorporating too many structure priors, we propose to implement a information-theoretic metric to measure the contribution of each structure prior to the localization, and hereafter, to integrate a most useful subset of all structure priors.
	To summarize, this paper makes the following contributions:
	\begin{enumerate}
		\item Firstly, we extend our previous work \cite{lyu2022structure} and further relief the feature deficiency problem by integrating 3D point, line, and plane features from various sensor modalities to relief the feature deficiency in civil environments. The heterogeneous features based localization is modeled in a sliding window fashion and solved with a graph optimization method.
		
		\item 
		With the heterogeneous feature used, we integrate more generic and broader range of prior information which we named as \textit{structure priors} and parameterized as the relative distances/angles between different geometric primitives.
		The association between observation and map is therefore simplified to one dimension level.
		To ease the burden of integrating too many factors, we develop a structure prior information selection strategy based on the \textit{information gain} to incorporate only the most effective structure priors for localization.
		\item Finally, we test our proposed framework extensively based on synthetic data, public datasets, and real UAV flight data obtained from both indoor and outdoor inspection tasks. The results indicate that the proposed framework can improve the localization robustness  even in challenging environments.
	\end{enumerate} 
	
	The remainder of the paper is organized as follows. In Section \ref{related_works}, the related works are provided. The proposed SPINS framework is formulated in Section \ref{system_description}. The geometric features and structure priors are modeled in Section \ref{features} and Section \ref{priors}, respectively. Experiment validations are provided in Section \ref{experiment}. Section \ref{conclusion} concludes the paper.
	\section{Related works}
	\label{related_works}
	SLAM-based localization is considered as one of the most promising approaches for robot localization. By measuring salient features from the environment, the robot can accumulatively estimate its pose in local coordinates, which is preferred in many challenging environments, such as complex indoor or urban cities \cite{weiss2011monocular}. In this part, we review the recent SLAM results based on different features and structure priors.
	
	Among existing methods reported in the literature, the point feature is most commonly used in the environment perception front-end in SLAM frameworks, especially in the visual aided navigation frameworks. Some successful demonstrations, such as ORB-SLAM \cite{Mur2015ORB}, and VINS-mono \cite{Qin2018VINSMono}, utilize 2D point features from vision sensors to estimate the local motion and sparse 3D point clouds of the environment. With the development of more advanced 3D sensing technologies, navigation methods based on 3D sensors, such as stereo-vision\cite{lemaire2007vision}, LiDAR\cite{zhang2014loam}, and RGB-D camera\cite{sturm2012benchmark}, can directly utilize 3D points with metric information in the estimation process, therefore can provide improved localization and reconstruction results.
	In addition to the most commonly used point features, line and plane features are also adopted as additional features in some mission scenarios, such as indoor servicing \cite{Padhy2019Monocular,Lu_2015_ICCV}, structure inspection \cite{hasan2016construction,nguyen2020liro}, and autonomous landing \cite{Andres2017Homography}, in case that point features are not sufficient. 
	
	In the past few years, SLAM methods using heterogeneous features have begun to draw researchers' attention. 
	The improvement of localization performance has been verified by works with different feature combinations. In  vision-based SLAM, line features are considered more effective than point features in a low textural but high structural environment with a proper definition of the state and re-projection error. 
	Extended from the ORB-SLAM, the PL-SLAM \cite{pumarola2017pl} can simultaneously handle both point and line correspondences. The line state is parameterized with its endpoints, and the re-projection error is defined as point-to-plane distances between the projected endpoints of 3D line and the observed line on image plane. 
	A tightly coupled Visual Inertial Odometry (VIO) exploiting both point and line features is proposed in \cite{he2018pl}. The line is parameterized as a six-parameter Pl{\"u}cker coordinate\cite{hodge1994methods} for transformation and projection simplicity, and a four-parameter orthonormal representation for optimization compactness. Similar reprojection error to \cite{pumarola2017pl} is utilized in \cite{he2018pl,lyu2022structure}. Comparisons of different line feature parameterization are provided in \cite{Yang2019Visual} based on the MSCKF SLAM framework, which shows that the closest point (CP) based \cite{yang2019aided} and quaternion based \cite{kottas2013efficient} representations outperform the Pl{\"u}cker representation under noisy measurement conditions.
	Besides the monocular based methods above, a stereovision-based VIO using point and line features together is proposed in \cite{zheng2018trifo} where the measurement model is directly extended from the monocular camera model similar to \cite{pumarola2017pl}. In the stereo vision based PL-SLAM framework \cite{gomez2019pl}, a visual odometry is formulated similarly to \cite{zheng2018trifo}. In addition to that, the key-frame selection and loop closure detection under point and line features setup are also provided. 
	
	As 3D sensors such as LiDAR or RGB-D camera become more popular, plane features now can be effectively extracted in a man-made environment. A Lidar Odometry and Mapping (LOAM) in real-time is proposed in \cite{zhang2014loam}, which utilizes plane features to improve the registration accuracy of a point cloud. A LiDAR-inertial SLAM framework based on 3D planes is proposed in \cite{geneva2018lips}, where the closest point representation is utilized for parameterizing a plane. A tightly-coupled vision-aided inertial navigation framework combining point and plane features is proposed in \cite{yang2019tightly}, where a plane is parameterized similar to \cite{geneva2018lips}. In addition, the point-on-plane constraint is incorporated to improve the VINS performance. Recently, point, line, and plane features are jointly applied in the SLAM frameworks \cite{zhang2019structure,yang2019observability,li2020leveraging,Aloise2019Systematic} to realize stable localization and mapping in low texture environments. In \cite{zhang2019structure}, the line and plane features are tracked simultaneously along a long distance to provide persistent measurements. Also, the relationship between features, such as co-planar points, is implemented to enforce structure awareness. Similar work \cite{li2020leveraging} utilizes line and plane to improve the feature richness and incorporates more spatial constraints to realize more robust visual inertial odometry. A pose-landmark graph optimization back-end is proposed in \cite{Aloise2019Systematic} based on the three types of features, which are handled in a unified manner in \cite{Nardi2019Unified}.
	A thorough theoretical analysis of implementing point, line, and plane features in VINS is provided in \cite{Yang2019Visual} where three kinds of features are parameterized as measurements to estimate the local state based on a recursive MSCKF framework. More importantly, the observability analysis of different combinations of features is provided, and the effect of degenerate motion is studied. 
	
	Although researchers have begun to introduce heterogeneous geometric features into their SLAM works, integrating prior map information to the localization still draws limited attention. There are mainly two types of prior information that can be obtained from a prior map to aid the SLAM, namely the global information and the local structure information. 
	By incorporating global pose constraints by matching local observations with a consistent global map, the SLAM drift can be reduced. Inspiring by this, the global information is incorporated in the local SLAM in both loosely coupled manner \cite{platinsky2020collaborative} and  tightly coupled manner \cite{middelberg2014scalable}.
	
	On the other hand, only limited structure priors information, such as point-on-line, line-on-plane, and point-on-plane constraints, are considered in the SLAM. A visual inertial navigation system that utilize both point features and actively selected line features on image plane is proposed in \cite{lyu2022structure}, structure prior information based on special point and line relationships are utilized to improve the localization performance. Note that the structure information is ubiquitous in civil environments that are rich of man-made objects. It is a practical wisdom to implement more general structure prior information of the environment to improve the localization and mapping quality rather than to consider the environment as entirely unknown. For instance, in a building inspection environment, the structure information, as elaborate as the CAD modes, or as coarse as some common sense such as flat planes, parallel lines, with proper parameterization, can be implemented to aid the localization process\cite{hasan2016construction,Jovan2016Matching}.
	
	\section{System description}
	\label{system_description}
\begin{figure}
		\centering
		\includegraphics[width=0.75\linewidth]{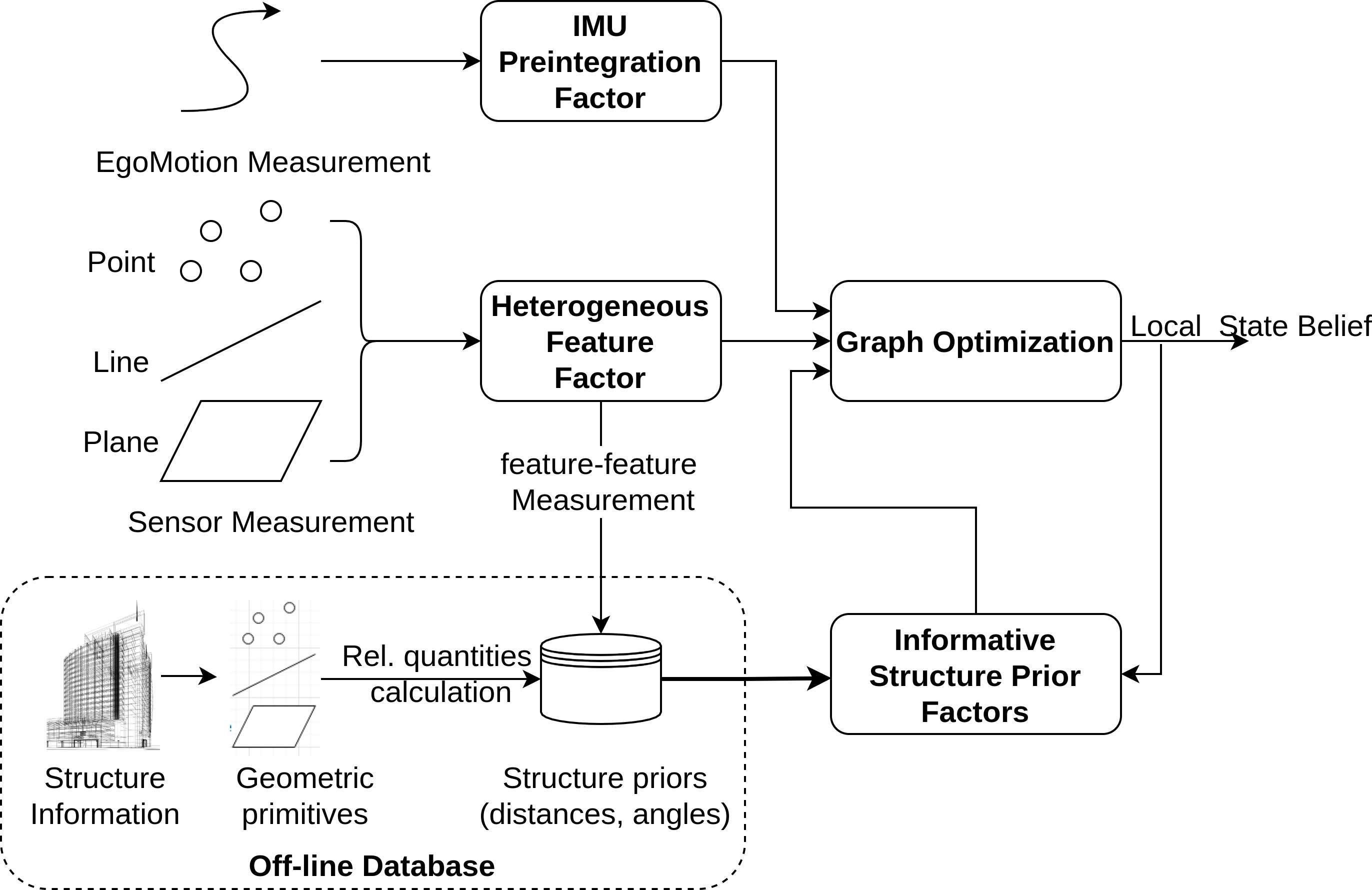}
		\caption{The systematic functional blocks of the proposed SPINS framework.}
		\label{fig:system-flowchart}
	\end{figure}

	In this section, the SPINS is described from a systematic point of view. The functional blocks of the system are illustrated in Fig. \ref{fig:system-flowchart}. 
	The SPINS depends on three types of information to fulfill the task of accurately and reliably localizing an autonomous vehicle in a challenging civilian environment, namely 1)  ego-motion measurements from interoceptive sensors, such as the Inertial Measurement Unit (IMU), at high-frequency feeding streaming, and 2) detected/tracked point, line, and plane features from exteroceptive sensors, and 3) structure priors which are parameterized as pairwise high fidelity measurements between features.
	\begin{figure}
		\centering     %%% not \center
		\subfigure[A geometric representation of point, line and plane features.]{\label{fig:sub-first}\includegraphics[width=0.5\linewidth]{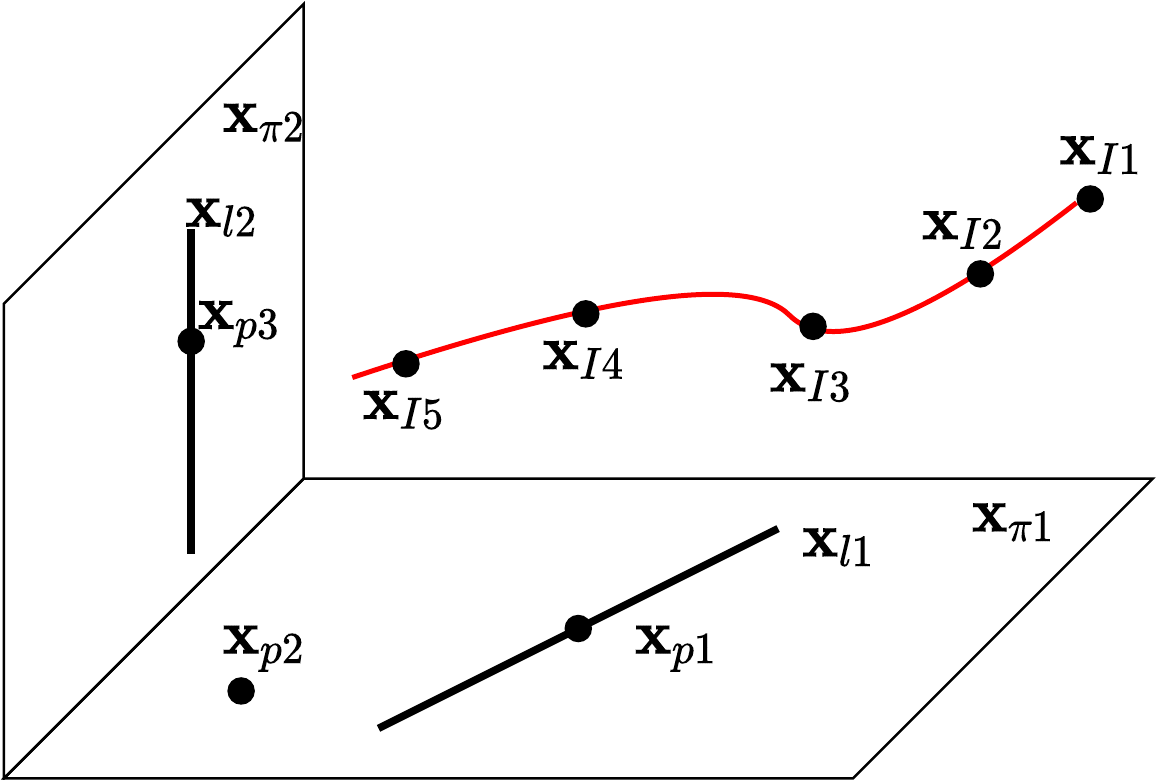}}
		\subfigure[The corresponding factor graph of subfigure (a).]{\label{fig:sub-second}\includegraphics[width=0.5\linewidth]{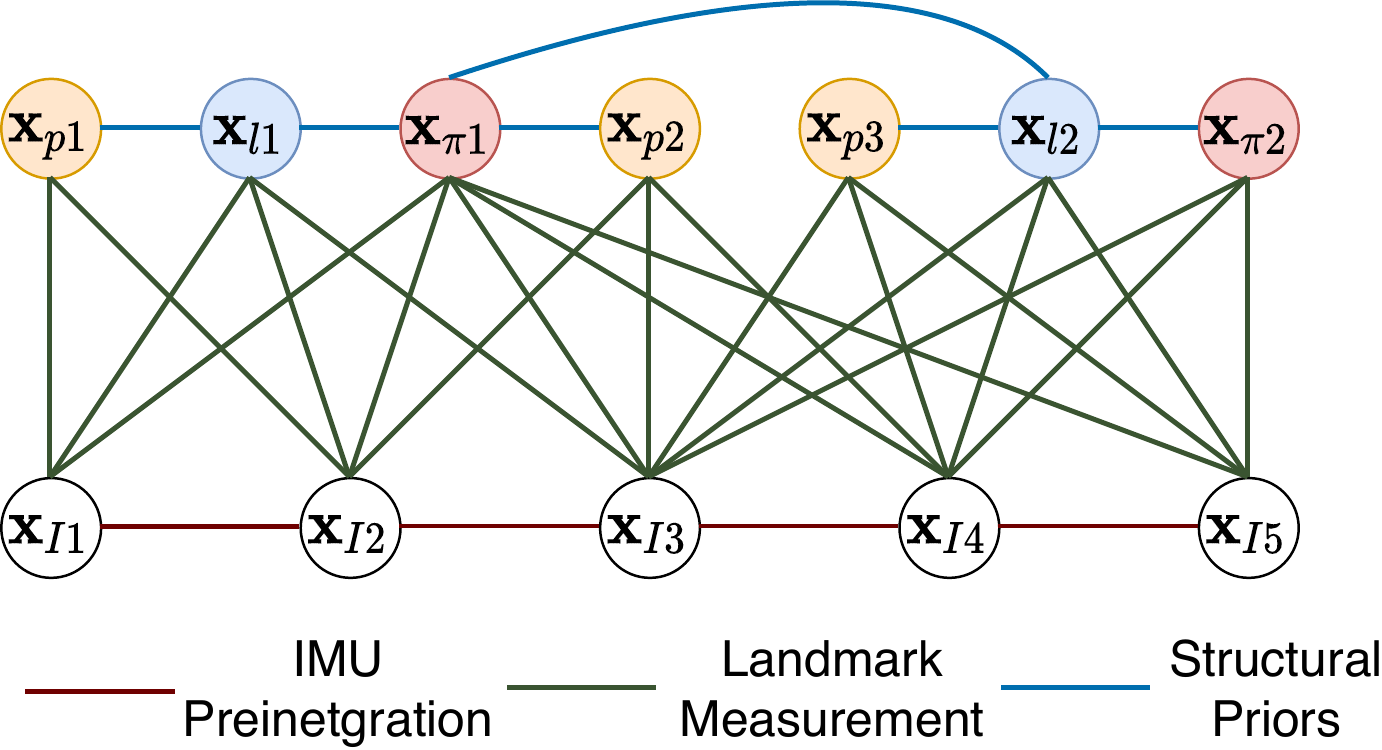}}
		\caption{The factor graph including the structure priors.}
	\end{figure}
	\subsection{Optimization formulation}
	The aforementioned three types of information within a sliding window are incorporated into a factor graph, which is a bipartite graph contains variables and factors. Specifically, the variables represent the state of the local vehicle and the heterogeneous features, and factors encode different types of observations and structure priors.
	The states included in the sliding window at time $t$ are defined as 
	\begin{equation}
		\label{state_def}
		\mathcal{X}_t \triangleq
		\begin{bmatrix}
			\{\xbf^\top_{I,m}\}_{m\in T_t} & \{\xbf^\top_{i}\}_{i\in P_t} &\{\xbf^\top_{j}\}_{j\in L_t} &\{\xbf^\top_{k}\}_{k\in \Pi_t}
		\end{bmatrix}^\top,
	\end{equation}
	where $\{\xbf_{I,m}\}_{m\in T_t}$ contains active IMU states within the sliding window at time instance $t$. $T_t$ denotes the set of IMU measurements at $t$. $P_t, L_t, \Pi_t$ denote the sets of point, line and plane features that are observed within the sliding window at time $t$, respectively. 
	The IMU state is 
	\begin{equation}
		\xbf_I\triangleq \begin{bmatrix}
			{^I_G}\bar{q}^\top & {^G}\pbf^\top_I & {^G}\vbf^\top_I & \bbf_g^\top & \bbf_a^\top 
		\end{bmatrix}^\top,
	\end{equation}
	where ${^I_G}\bar{q}$ is a unit quaternion denoting the rotation from the global frame $\{G\}$ to the IMU frame $\{I\}$. ${^G}\pbf_I$ and  ${^G}\vbf_I$ are the IMU position and velocity, respectively. $\bbf_g$, $\bbf_a$ are the random walk biases for gyroscope and accelerator, respectively. 
	With the state definition (\ref{state_def}), the objective is to minimize the cost function of different measurement residuals in (\ref{ful_cost}).
	\begin{equation}
		\begin{aligned}
			\label{ful_cost}
			\min\limits_{\mathcal{X}_t}
			&\left\{\|\rbf_{p}\|^2_{\Pbf_{p_t}} + \sum_{m\in T_t}\|\rbf_{I,m}\|^2_{\Sigma_m}\right.\\
			&+ \sum_{i\in P_t}\rho(\|\rbf_{i}\|^2_{\Sigma_i}) + \sum_{j\in L_t}\rho(\|\rbf_{j}\|^2_{\Sigma_j}) + \sum_{k\in \Pi_t}\rho(\|\rbf_{k}\|_{\Sigma_k}^2) \\
			&\left.+ \sum_{s\in \mathcal{S}_t}\rho(\|\rbf_s\|_{\Sigma_{s}}^2)\right\}.
		\end{aligned}
	\end{equation}
	The first term of (\ref{ful_cost}) is the cost on prior estimation residuals, and $\Pbf_{p_t}$ is the corresponding covariance prior to the optimization at $t$\cite{kaess2012isam2}. 
	The second term is the cost of IMU-based residual, and $\rbf_{I,m}$ defines the measurement residual between active frames $m$ and $m+1$. The IMU measurement between time step $m$ and $m+1$ is obtained by integrating high-frequency raw IMU measurements continuously with the technique called IMU preintegration \cite{Forster2017Manifold}, and $\Sigma_m$ is the corresponding measurement covariance.
	The second line of (\ref{ful_cost}) represents the cost function of measurement residuals of point, line, and plane features and are weighted by their corresponding covariances. The third line is the cost of structure priors measurement residuals.
	
	For a measurement in Euclidean space, its residual term is defined as the difference between the predicted measurement based on estimated state $\hat{\mathcal{X}}$ and a real measurement $\zbf$ , as 
	\begin{equation}
		\rbf = h(\hat{\mathcal{X}}) - \zbf,
	\end{equation}
	where $h(\cdot)$ is the measurement prediction function for the estimated state between any two variables in the factor graph. The term $||\rbf||^2_{\Sigma} = \rbf^\top\Sigma^{-1}\rbf$ is defined as the squared Mahalanobis distance with covariance matrix $\Sigma$. A huber loss $\rho(\cdot)$ \cite{huber1992robust} is applied on each squared term to reduce potential mismatches between states and measurements. 
	The optimization of $(\ref{ful_cost})$ is usually solved with an iterative Least-Squares solver through a linear approximation as detailed in Appendix \ref{solver}. 

	The formulation of measurement functions $h(\cdot)$ with regard to the geometric features and the structure priors are provided in \ref{features} and Section \ref{priors}, respectively.
	\subsection{Structure prior information}
	The structure information is ubiquitous, ranging from the fine-grained blueprint to common knowledge such as parallel lines or planes. The greatest challenge to integrate the prior information in the optimization is to correctly associate the structure information with the observed features. Benefiting from using point, line, and plane features simultaneously, we can parameterize the spatial relationship between different geometric features as pairwise relative distances and angles in Section \ref{priors}. The advantages of using such parameterization are mainly three folds. 
	\begin{enumerate} 
		\item First, the prior knowledge can be integrated in a simple fashion. With the distance and angle based formulation, the rigorous association process between the prior knowledge and current observation, which are normally based on high-dimensional and computational-demanding feature matching processes, can be simplified to a scalar level association based on thresholds.
		\item Second, more extensive prior knowledge can be utilized to aid the navigation. The distances and angles can not only be extracted from prior maps with specific format, but also be obtained from structural common senses, hand-measured quantities, and so on.
		\item As the angles and distances are stored as scalars, the storage can be reduced dramatically comparing to storing a map.
	\end{enumerate}
	
	The structure priors can be extracted offline based on the following three steps:
	\begin{enumerate}
		\item extract structural primitives from various formates, \eg high fidelity maps, local measurements, semantic information \etc,
		\item measure and calculate the relative quantities between every two primitives (as formulated in Section \ref{priors}), and
		\item store salient structure quantities (distances/angles) as structure priors in a database.
	\end{enumerate}
	{\begin{figure}
			\centering
			\includegraphics[width=0.75\linewidth]{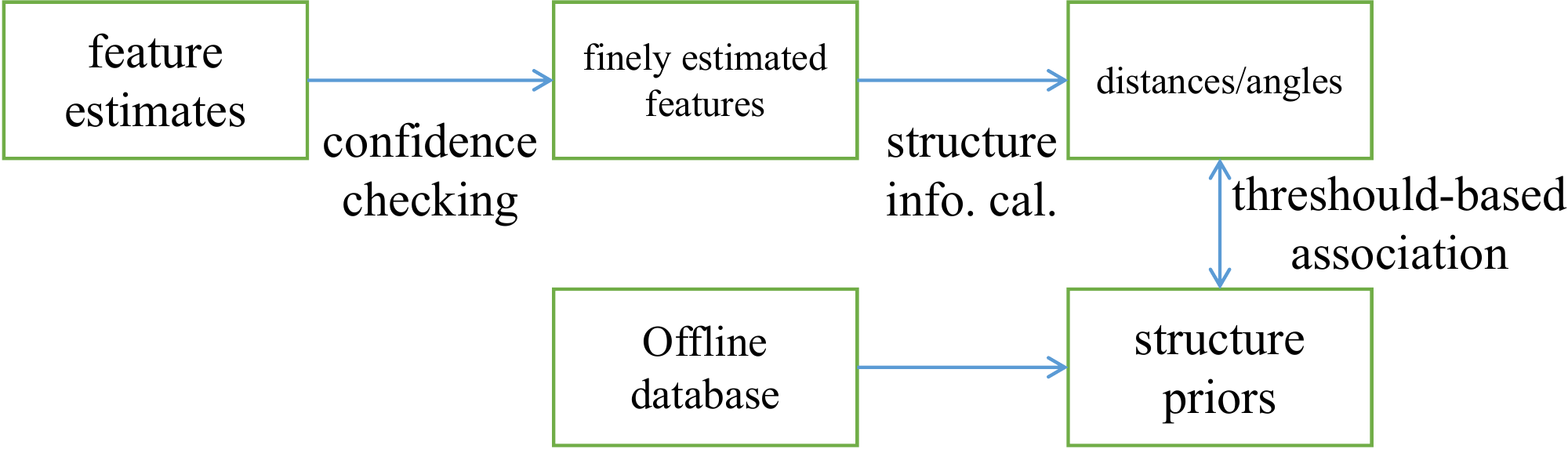}
			\caption{The online association between structure priors and local estimates.}
			\label{fig:association}
	\end{figure}}
	
	{ 
		The association process between offline structure prior information and online local observations is illustrated as Fig.\ref{fig:association}.	Initially, finely estimated features  within current optimization window are selected based on the estimation confidences. Then the structural quantities (angles/distances) based on the estimates can be calculated. Finally, a structure prior is integrated into the optimization when it is close enough to a structural quantity from estimates, based on a scalar threshold. 
		{ Given two features' estimates, $\hat{\xbf}_i,\hat{\xbf}_j$, a distance/angle prior information $z_s$ is associated to $i,j$ when 
			\begin{equation*}
				|h_s(\hat{\xbf}_i,\hat{\xbf}_j) - z_s|\le \tau_s,
			\end{equation*}
			where $h_s(\cdot)$ is the function to extract distance/angles (see Sec. V), and $\tau_s$ is a scalar threshold. To prevent possible false association of structure prior information, we set a very small threshold $\tau_s$, and allow structure priors to be assigned to only finely estimated features.}
		
		Above structure priors integration process should be carried out based on following pre-requisites.
		First, the structure priors should be quantities that are identifiable among all relative distances and angles. Second, the structure prior quantity should be representative of the most salient structural patterns of an environment. In many robotic operation environment, such as indoor navigation, building inspection, geometric features are ubiquitous, and structure patterns are inerratic and repetitive. In such environment, the extracted structure priors (angles and distances) are sparsely distributed, see, e.g. Fig.\ref{fig:room_planes-crop} and Fig.\ref{fig:distributions}, and the association can be achieved based on a simple threshold as described above. In this paper, we only consider structure rich environments where the sparseness of the structure priors holds.
	}
	
	\section{Heterogeneous geometric features}
	\label{features}
	In this section, we discuss how to model the point, line, and plane features based on on-board perception and incorporate them into the graph optimization.
	\begin{figure}
		\centering
		\includegraphics[width=0.75\linewidth]{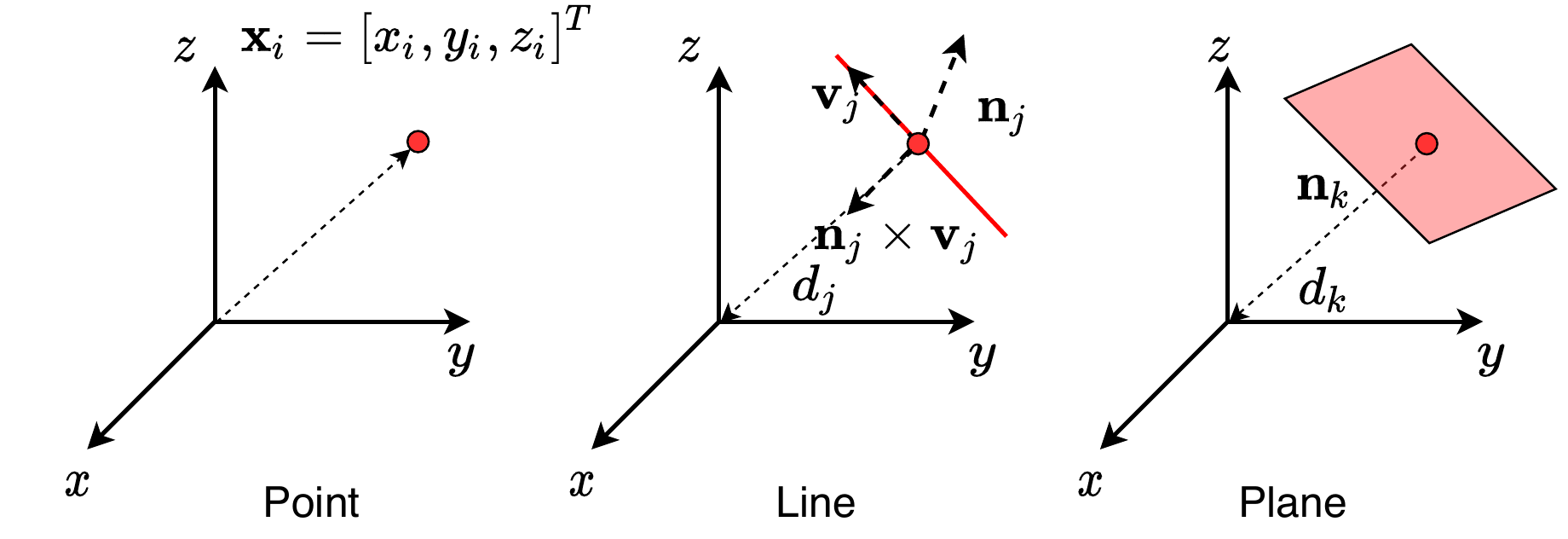}
		\caption{The geometry representation of point, line and plane in 3D space.}
		\label{fig:geometry_point}
	\end{figure}
	\subsection{Point feature}
	As one of the most frequently implemented features in perception tasks, a point feature can be uniquely parameterized by its 3D coordinate, as shown in Fig. \ref{fig:geometry_point}.
	A point feature $i\in P$ can be extracted from different sensors by measuring it in the local frame,
	\begin{equation}
		\zbf_{i} = h_i(\xbf_I, \xbf_i)+\pmb\nu_i = {^{I}_{G}}\Rbf ({^G}\pbf_i - {^G}\pbf_I)+ \pmb\nu_i,
	\end{equation}
	where $^G\pbf_i\in \mathbb{R}^3$ and $^G\pbf_I\in \mathbb{R}^3$ are the 3D positions of the point feature and the local vehicle in the global frame. $\pmb\nu_i\in \mathbb{R}^3$ is the measurement noise. ${^{I}_{G}}\Rbf\in \mathbb{SO}(3)$ represents the rotation from the local frame to the global frame. By defining the state of point feature as $\xbf_i = {^G}\pbf_i$, we have the Jacobians calculated in Appendix \ref{point_feature}.
	
	\subsection{Line feature}
	One most commonly implemented representation of a line feature in 3D space is its Pl\"ucker coordinates. For an infinite line $j\in L$, its Pl\"ucker coordinates are defined as $\lbf_j \triangleq \begin{bmatrix}
		\nbf_j^\top & \vbf_j^\top
	\end{bmatrix}^\top$, $\nbf_j$ and $\vbf_j$ are the normal vector and directional vector calculated from any two distinct points on the line, as shown in Fig. \ref{fig:geometry_point}.
	
	A measurement of an infinite line can be modeled as its Pl\"ucker coordinates in the local frame as 
	\begin{equation}
		\begin{aligned}
			\zbf_j &= h_j( {^G}\xbf_I, {^G}\lbf_j ) + \pmb\nu_j \\& = \left[\begin{array}{cc}
				_{G}^{I}\Rbf & -_{G}^{I} \mathbf{R}\left[^{G} \pbf_{I} \right]_{\times} \\
				\mathbf{0}_{3} & {_{G}^{I}}\mathbf{R}
			\end{array}\right] {^G}{\mathbf{l}}_j + \pmb\nu_j, 
		\end{aligned}
	\end{equation} 
	where $\pmb\nu_j\in \mathbb{R}^6$ is the measurement noise. 
	
	Obviously, the expression $\lbf_j$ is not a minimum parameterization of the line state. To calculate the Jacobian, we implement the closest point approach described in \cite{yang2019observability}, which is formulated as 
	$\pbf_j = d_j\bar q_j\in \mathbb{R}^4$, 
	where the unit quaternion $\bar{q}_j$ and the closest distance of the line to the origin can be calculated from the Pl\"ucker coordinates respectively as 
	\begin{align}
		\Rbf_j(\bar{q}_j) &= \begin{bmatrix}
			\frac{\nbf_j}{\|\nbf_j\|} & \frac{\vbf_j}{\|\vbf_j\|} & \frac{\nbf_j}{\|\nbf_j\|}\times \frac{\vbf_j}{\|\vbf_j\|}
		\end{bmatrix},\\
		d_j &= \frac{\|\nbf_j\|}{\|\vbf_j\|},
	\end{align}
	where $\Rbf_j(\bar{q}_j)$ is the corresponding rotation matrix to $\bar{q}_j$.
	By defining the line state as $\xbf_j = {^G}\pbf_j$, we have a minimum parameterization of the line in Euclidean space. The corresponding measurement Jacobians are provided in Appendix \ref{line_feature}.
	
	\subsection{Plane feature}
	An infinite plane $k\in \Pi$ can be minimally parameterized by the closest point $\pbf_k = d_k\nbf_k \in \mathbb{R}^3$, as shown in Fig. \ref{fig:geometry_point},
	where $\nbf_{k} $ is the plane's unit normal vector, and $d_{k} $ is the distance from the origin to the plane. The plane measurement here is modeled as the closest point in the local frame as 
	\begin{equation}
		\label{plane_meas}
		\begin{aligned}
			\zbf_{k} &= {^I}\pbf_k + \pmb\nu_k = {^I}\nbf_{k}{^I}d_{k} + \pmb\nu_k,
		\end{aligned}
	\end{equation}
	where $\pmb\nu_k\in \mathbb{R}^3$ represents the plane measurement noise.
	The translation of the unit normal vector and the distance of a plane from the global frame to the local frame is 
	\begin{equation}
		\label{plane_trans}
		\begin{bmatrix}
			^I\nbf_{k}\\^I d_{k}
		\end{bmatrix} = \begin{bmatrix}
			{^I_G}\Rbf & \mathbf{0}_{3\times1}\\ -^G\pbf_I^\top &1
		\end{bmatrix}\begin{bmatrix}
			^G\nbf_{k}\\^G d_{k}
		\end{bmatrix}.
	\end{equation}
	Incorporating (\ref{plane_trans}) into (\ref{plane_meas}), the plane measurement can be expressed with the normal vector $^G\pbf_I^T$ and distance $^G d_{\pi}$ in the global frame as 
	\begin{equation}
		\label{plane_GtoI}
		\begin{aligned}
			\zbf_k = {^I} d_{k} {^I}\nbf_{k}+\pmb\nu_{k} = (-^G\pbf_I^\top {^G}\nbf_{k} + {^G} d_{k}){^I_G}\Rbf^G\nbf_{k}+\pmb\nu_{k}\\
			=-^G\nbf_{k}^\top {^G}\pbf_I{^I_G}\Rbf^G\nbf_{k} + {^G} d_{k}{^I_G}\Rbf{^G}\nbf_{k}+\pmb\nu_{k}.
		\end{aligned}
	\end{equation}
	Define the state of a plane $k$ as $\xbf_k={^G}\pbf_k={^G}d_k{^G}\nbf_k$, which is the closest point of the plane to the origin. we can calculate the measurement Jacobian in Appendix \ref{plane_feature}.
	
	\begin{remark}
		In our paper, the raw sensor measurement models of different geometric features are not explicitly provided since they are highly dependent on the sensing mechanism of different sensors. Our purpose here is to provide a common framework implementing the heterogeneous features rather than considering a specific sensor.
	\end{remark}
	
	\section{Structure priors formulation}
	\label{priors}
	In this part, the structure priors are formulated as the relative relationships between features. Specifically, the angles and distances are defined between different geometric primitives, including point, line, and plane.
	\subsection{Feature-to-feature prior modeling}
	The structure prior factors are plotted as blue edges in the factor graph, as shown in Fig. \ref{fig:factor_graph}.
	Denote the topology set containing all the pairwise structure priors as $\mathcal{S}$, then an edge $(a,b)\in \mathcal{S}$ indicates that some quantitative measurements between two features $a,b\in \{P,L,\Pi\}$, denoted as $\zbf_{ab}$, are known a priori. 
	Let $h_{ab}$ denote the measurement function between $a$ and $b$, the structure prior residual can be obtained as
	\begin{equation}
		\rbf_{ab} = h_{ab}(\hat\xbf_a, \hat\xbf_b) - \zbf_{ab}.
	\end{equation}
	The residual cost is
	\begin{equation}
		\sum\limits_{(a,b)\in \mathcal{S}}\|\rbf_{ab}\|^2_{\Sigma_{ab}},
	\end{equation}
	where $\Sigma_{ab}$ is the covariance of the measurement noise which represents the fidelity of implementing specific structure prior constraints. The covariance is assumed to follow Gaussian distribution and is calculated statistically during the structure priors extraction process.
	The following are to model the pairwise measurements between point, line, and plane features described above.
	\subsection{Feature-to-feature factors}
	\subsubsection{Point-to-point factor}
	When two salient points are detected, the possible structure prior information that characterizes the spatial relationship can be modeled as a $1-3$ dimensional measurement. 
	Denote two point features $i,i'\in P$, and their relative translation $\xbf_{ii'} =  \xbf_{i'} -  \xbf_i$, the point-to-point structure measurement
	\begin{equation}
		\zbf_{ii'} = h_{ii'}(\xbf_{ii'}),
	\end{equation}
	is to project the 3D displacement between the two points onto a specific 1-3 dimensional metric in the global frame. 
	Specifically, the distance measurement can be modeled as $z_{ii'}^d = \|\xbf_{ii'}\|$. 
	
	The measurement residual Jacobians with respect to the points state are provided in Appendix \ref{ptpj}. 
	
	\begin{remark}
		In the point-to-point structure, the points should be salient in both texture and structure senses. In practice, most points are distributed according to the texture, and it may not be easy to endow structure information. Some examples of structurally salient points are intersection points, endpoints, and corner points. The integration of point-to-point structure prior information depends on the extraction and recognition of structural points, which may be challenging in practice. 
	\end{remark}
	\subsubsection{Point-to-line factor}
	The spatial relationship between a point and an infinite line can be described with a 2D vector. With a point $i\in P$ and a line $j\in L$, we define a 2D displacement between them as 
	\begin{equation}
		\xbf_{ij} =\begin{bmatrix}
			\bar\nbf_j^\top \\ \bar\nbf_j^\top\times\bar\vbf_j^\top 
		\end{bmatrix} \xbf_i + \begin{bmatrix}
			0 \\d_j
		\end{bmatrix}\in \mathbb{R}^2,
	\end{equation}
	where $\bar\nbf_j = \frac{\nbf_j}{\|\nbf_j\|}$, $\bar\vbf_j = \frac{\vbf_j}{\|\vbf_j\|}$, and $d_j$ are the line $j$'s unit normal vector, unit directional vector, and the distance to the origin point, respectively.
	Denote the point-to-line measurement of $\xbf_{ij}$ as
	\begin{equation}
		\zbf_{ij} = h_{ij} ( \xbf_{ij}),
	\end{equation}
	The point to line distance can be calculated by letting $h_{ij}(\cdot)$ be a norm operator, 
	\begin{equation}
		z_{ij}^d=d_{ij} = \|\xbf_{ij}\|.
	\end{equation}
	Hereafter, the point-on-line constraint can be enforced as $ d_{ij}= 0$. The measurement residual Jacobian is calculated as Appendix \ref{ptlj}.
	\subsubsection{Point-to-plane factor}
	The relationship between a point and an infinite plane can be described with one scalar, \ie the distance from the point to the plane.
	With a point feature $i\in P$, and an infinite plane feature $k\in \Pi$, the distance between a point and a plane is defined as 
	\begin{equation}
		d_{ik} = (\nbf_k)^\top\pbf_i + d_k.
	\end{equation}
	Define a measurement function as $z^d_{ik} =  d_{ik}$,
	then the point-on-plane constraint can be enforced by letting $ d_{ik}= 0$. The measurement residual Jacobian is provided in Appendix \ref{ptplj}.
	
	\subsubsection{Line-to-line factor}
	The relationship of two lines can be uniquely parameterized with a 3D translation vector and a rotation angle. 	
	Given two lines, denoted respectively as $j, j'\in L$, the rotation $\alpha_{jj'}$ and translation $\dbf_{jj'}\in \mathbb{R}^3$ can be calculated as follows: 
	\begin{equation}
		\alpha_{jj'} = \vbf_j^\top\vbf_{j'},
	\end{equation}
	and 
	\begin{small}
		\begin{equation}
			\dbf_{jj'} = \left\{\begin{array}{ll}
				\mathbf{0}, & j\text{ and } j' \text{ intersection},\\
				\bar\dbf_{jj'}, & j\text{ and }j' \text{ are parallel},\\
				(\bar\vbf_j\times\bar\vbf_j')^\top\bar\dbf_{jj'}(\bar\vbf_j\times\bar\vbf_j'), & \text{otherwise},
			\end{array}\right.
		\end{equation}
	\end{small}
	where $\bar \dbf_{jj'} = d_j\bar\nbf_j\times\bar\vbf_j - d_{j'}\bar\nbf_{j'}\times\bar\vbf_{j'}$. 
	$d_j\bar\nbf_j\times\bar\vbf_j$ and $d_{j'}\bar\nbf_{j'}\times\bar\vbf_{j'}$ are the closest point of line $j$ and $j'$ to the origin. It is straightforward to prove that as the relative angle $\alpha_{jj'}\to \pm 1$, the distances $d_{jj'} = \|\dbf_{jj'} \|$ denotes the relative distance between two parallel lines.

	We first consider the rotation $\alpha_{jj'}$ as a measurement between two lines. Further, when two lines are parallel, namely $\alpha_{jj'} = \pm 1$, the distance $d_{jj'} = \|\dbf_{jj'}\|$ is considered as another measurement, namely 
	\begin{equation}
		\zbf_{jj'} = \left\{
		\begin{aligned}
			&\begin{bmatrix}
				&\alpha_{jj'} & d_{jj'}
			\end{bmatrix}^\top, & \text{in parallel},
			\\
			&\alpha_{jj'}, & \text{otherwise}.\end{aligned}\right.
	\end{equation}
	The Jacobian of line-line measurement residual is provided in \ref{ltlj}.
	
	\subsubsection{Line-to-plane factor}
	The spatial relationship between a line and a plane can be characterized by the dot product of the directional vector of a line $j\in L$ and the normal vector of a plane $k\in \Pi$, denoted as $\alpha_{jk}$: 
	\begin{equation}
		\alpha_{jk} = \bar\vbf_{j}^\top \nbf_k.
	\end{equation} 
	Especially, when $\alpha_{jk} = 0$, namely a line is parallel to a plane, a distance can be further calculated as 
	\begin{equation}
		d_{jk} = \nbf_k^T(\bar\nbf_j\times \bar\vbf_j)d_j - d_{k}.
	\end{equation}
	The measurement therefore is 
	\begin{equation}
		\zbf_{jk} = \left\{
		\begin{aligned}
			&\begin{bmatrix}
				\alpha_{jk} & d_{jk}
			\end{bmatrix}^\top, & \text{in parallel},\\
			&\alpha_{jk}, &\text{otherwise}.
		\end{aligned}\right.
	\end{equation}
	Specifically, the line-on-plane constraint is enforced as $\alpha_{jk} =0$, and $d_{jk} = 0$.
	The associated Jacobian is provided in Appendix \ref{ltpj}.
	
	\subsubsection{Plane-to-plane factor}
	Similar to the above formulations, a similar dot product between the unit normal vectors of two planes, can be calculated as 
	\begin{equation}
		\alpha_{kk'} = \nbf_k^\top\nbf_{k'}.
	\end{equation}
	When two planes are parallel, the displacement can be calculated based on the closest points of two planes as
	\begin{equation}
		\dbf_{kk'} = \xbf_{k'} -\xbf_k = \nbf_k (d_{k'} - d_{k})= \nbf_{k'} (d_{k'} - d_{k}).
	\end{equation}
	Define the measurement of the relationship as 
	\begin{equation}
		\zbf_{jk} = \left\{
		\begin{aligned}
			&\begin{bmatrix}
				\alpha_{kk'} & d_{kk'}
			\end{bmatrix}^T, & k \text{ parallel to } k',\\
			&\alpha_{kk'}, &\text{otherwise}.
		\end{aligned}\right.
	\end{equation}
	The  measurement Jacobians are provided in Appendix \ref{pltplj}.
	\begin{table}[]
		\caption{The structure priors formulated as disances/angles} 
		\label{sp_table}
		\centering
		\begin{tabular}{|l|l|l|l|}
			\hline 
			& Point $i$ & Line $j$                                                                                                                                            & Plane $k$                                                                                                                                                                              \\ \hline\hline
			$i$ & $d_{ii'}$ & \begin{tabular}[c]{@{}l@{}}$d_{ij}$\\ e.g.\\ point-on-line: $d_{ij}=0$\end{tabular}                                                                 & \begin{tabular}[c]{@{}l@{}}$d_{ik}$\\ e.g.\\ point-on-plane: $d_{ik}=0$\end{tabular}                                                                                                    \\ \hline
			$j$  & -------  & \begin{tabular}[c]{@{}l@{}}$\alpha_{jj'}$\\ $d_{jj'}$ if parallel \\ e.g.\\ orthogonality: $\alpha_{ij} = \pm 0$\end{tabular} & \begin{tabular}[c]{@{}l@{}}$\alpha_{jk}$\\ $d_{jk}$ if parallel ($\alpha_{jk} = 0$)\\ e.g.\\ line-on-plane:\\$d_{jk}=0, \alpha_{jk}=0$\\ orthogonality: $\alpha_{jk}=\pm 1$\end{tabular} \\ \hline
			$k$ &   -------    &   -------                                                                                                                                                  & \begin{tabular}[c]{@{}l@{}}$\alpha_{kk'}$\\ $d_{kk'}$ if parallel $\alpha_{kk'}= \pm 1$\\ e.g.\\ orthogonality: $\alpha_{kk'} = \pm 1$\end{tabular}                                    \\ \hline 
		\end{tabular}
	\end{table}
	With the above formulation of spatial relationships between features, the structure priors can be encoded into low dimensional angles and/or distances, as summarized in Tab \ref{sp_table}. The low dimensional encoding makes their associations with the structure priors database easy. Both the heterogeneous geometric feature factors and the structure prior factors are further integrated with the graph optimization toolbox GTSAM \cite{dellaert2012factor}.
	
	\subsection{Structure prior selection}
	\label{structure_selection}
	\begin{figure}
		\centering
		\includegraphics[width=0.75\linewidth]{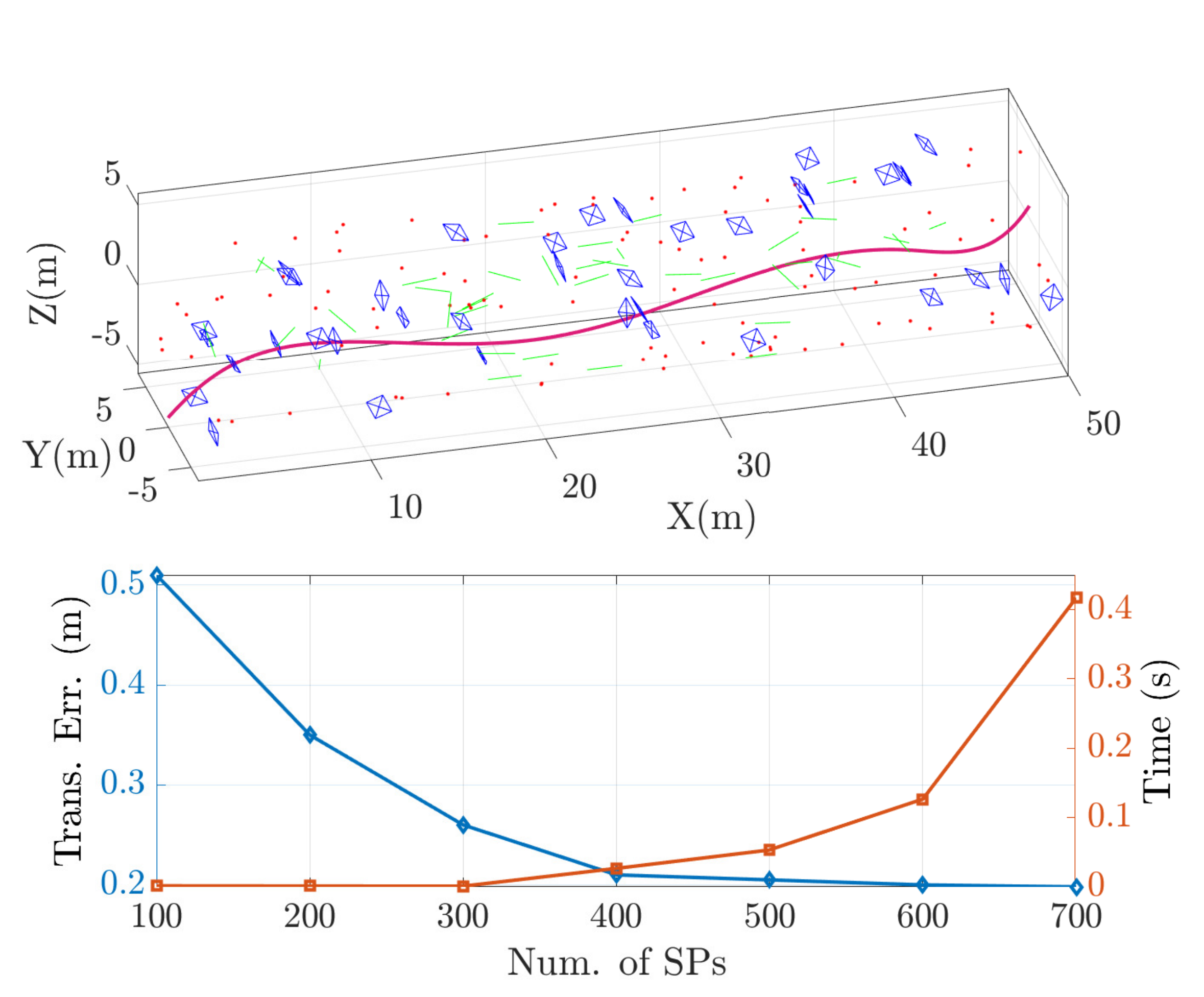}
		\caption{The translational estimation error and calculation time as number of SP grows.}
		\label{fig:trajs-crop}
	\end{figure}
	
	Based on the above formulations of the structure information between features, there are at most $n(n-1)/2$ possible feature priors in a scenario with $n$ geometric features. 
	In the graph optimization process, incorporating too many structure priors will severely damage the sparsity of the graph matrix, therefore will slow down the optimization. As an illustrative example, a synthetic environment with 100 points, 40 lines, and 40 planes are shown in Fig. \ref{fig:trajs-crop}. 
	Despite the localization error decreases as more structure priors are incorporated into the optimization, the optimization efficiency deteriorates simultaneously.
	Among all the potential prior information, some are not as helpful as others, and there may also exist redundancies in the structure priors set. 
	
	Based on above observations, it is a practical trick to select a limited number of structure prior measurements which benefit the localization the most.
	Specifically, we consider to minimize the localization uncertainty represented by a estimation covariance. 
	In this paper, we consider implementing the Fisher Information Matrix (FIM) to measure the contribution of a structure prior to the localization performance. Denote the belief of the state within the sliding window of time $t$ as $\mathcal{X}_t\sim \mathcal{N}(\bar{\mathcal{X}}_t, \Pbf_t)$, and one structure prior $s\in\mathcal{S}$ of current local map as a measurement, $ h_s(\mathcal{X}_t)\sim \mathcal{N}(\zbf_s, \Sigma_s)$, we have the following equation according to the Bayes' rule 
	\begin{equation}
		\label{bayes}
		\Pbf_{t^+}^{-1} = \Pbf_{t}^{-1} + \sum\limits_{s\in \mathcal{S}_t}\mathbf{I}_s,
	\end{equation}
	where $ \mathbf{I}_s = \Jbf_s^T\Sigma_s^{-1}\Jbf_s$ is the FIM of a specific structure prior $s$. $\Jbf_s$ is the Jacobian of the structure prior $s$ with regard to the local pose. $\Pbf_{t}$ and $\Pbf_{t^+}$ denote the covariance before and after integrating the structure priors, respectively.
	
	Finally, we can use the marginalization technique \cite{Carlone2019Attention} to obtain the localization uncertainty $\Pbf_t'$ by marginalizing out other states.
	The structure priors can be selected by minimizing a metric of the covariance  $\Pbf_t'$. 
	The selection problem is NP-hard and cannot be solved efficiently for a large number of structure priors.
	As indicated in \cite{shamaiah2010greedy}, the $\log\det(\cdot)$ metric of the covariance is submodular w.r.t. the information gain of structure priors. With efficient greedy algorithms, a sub-optimal solution with guaranteed performance can be obtained more efficiently. We use selection algorithm similar to  \cite{lyu2022structure} to obtain the structure prior set.
	
	\section{Experiment evaluation}
	\label{experiment}
	In this part, the proposed SPINS is tested based on synthetic data, the public datasets, and, most importantly, on real flight datasets that are collected with a UAV during indoor and outdoor inspection tasks.
	\subsection{Synthetic data}
\begin{figure}
		\centering
		\includegraphics[width=0.75\linewidth]{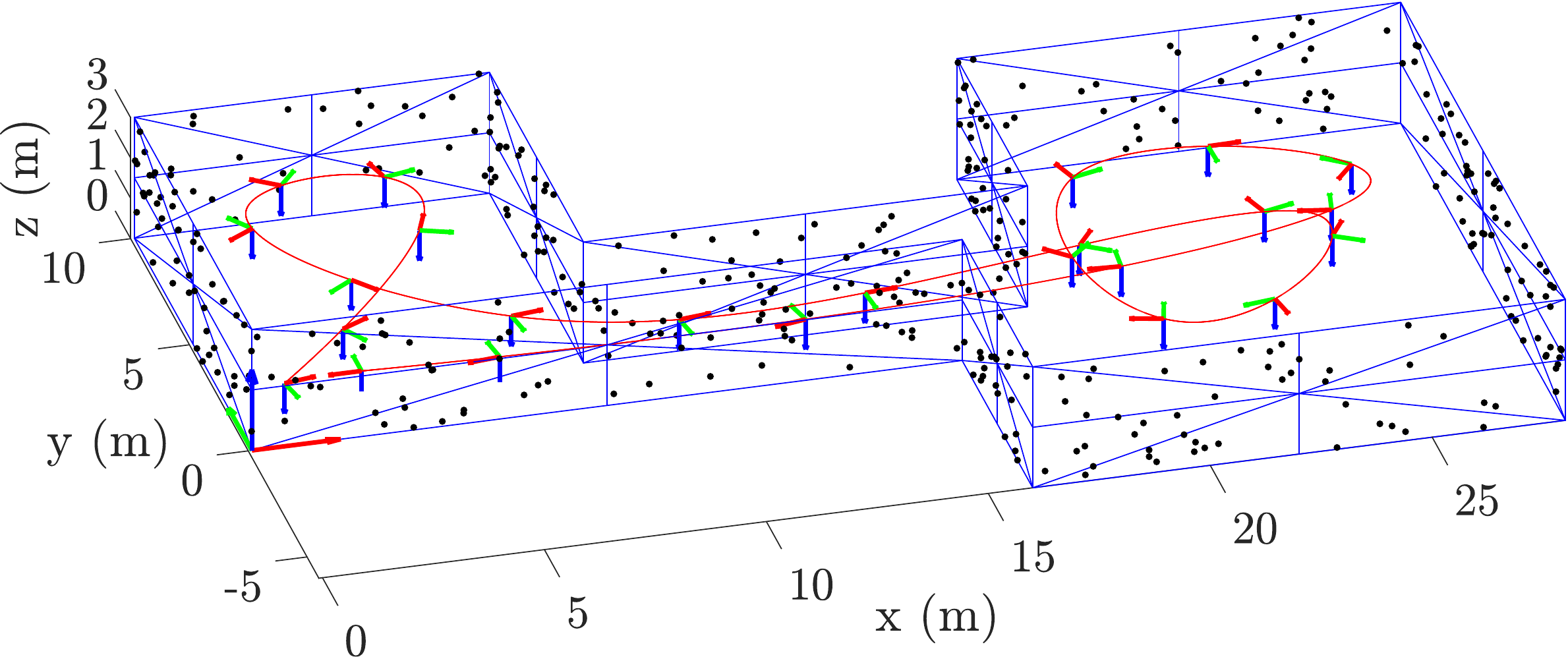}
		\caption{The 2.5D simulation scenario with point, line and plane features.}
		\label{fig:trajectory-crop}
	\end{figure}

	To evaluate the localization performance of the proposed framework, we create a customized 2.5D indoor simulation scenario with point, line, and plane features as presented in Fig. \ref{fig:trajectory-crop}. 
	A 3D robot trajectory is generated within the simulation space based on spline functions as a red curve. An IMU is simulated according to the ADIS16448 IMU sensor specifications listed in \cite{geneva2018lips}. We assume that the 3D geometry information of the features is obtained according to the measurement function described in Section \ref{features} with extra FOV limitations. 
	
	The optimization-based localization is solved with the iSAM2 solver from the GTSAM packag, which we additionally integrate line factors, plane factors, and structure prior factors. Comparisons based on different features and structure prior factors are carried out. Specifically, we consider 1) point feature (P-INS), 2) point and line features (PL-INS), 3) point, line, and plane features (PLP-INS) based methods, and 4) our structure prior aided method (SPINS). We select 1) 20 structure priors in each frame randomly (SPINS-Rand. 20 SP), 2)  20 structure priors according to Section \ref{structure_selection} (SPINS-App. Info. Opt. 20 SP). 3) the most informative 20 structure priors (SPINS-Info. Opt. 20 SP), and 4) all structure priors (SPINS-All SP). The structures information listed as Table \ref{sp} are adopted. { In addition, a prior map based localization method is used as a localization benchmark where perfect feature matching between local observations and map is assumed.}
	
\begin{figure}
		\centering
		\includegraphics[width=0.75\linewidth]{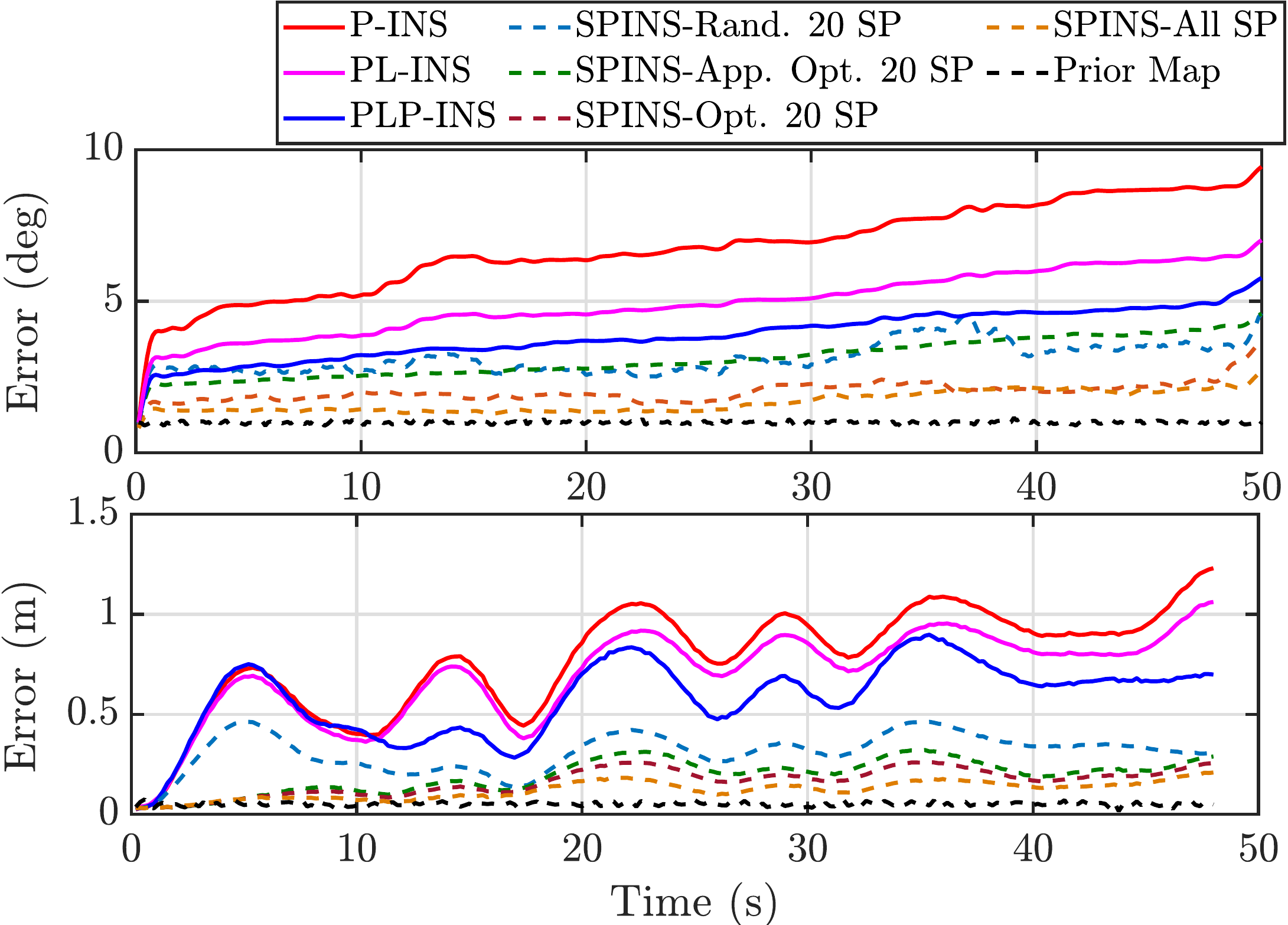}
		\caption{The translational and rotational RMSE under different feature and prior information configurations.}
		\label{fig:errors_sim-crop}
	\end{figure}

	The root-mean-square errors (RMSEs) are plotted in Fig. \ref{fig:errors_sim-crop}. The quantitative comparison between different strategies is also provided in Tab. \ref{sp}. 
	The localization results of using heterogeneous features are plotted as solid lines in different colors. It is apparent that, as more types of feature are used, more accurate estimation can be achieved. 
	More important, the integration of structure prior information can further improve the localization performance, as plotted in dashed lines. Specifically, integrating all structure prior information unsurprisingly achieves the closest localization performance to the prior map based method. Nevertheless, the computation time for solving each round of the local optimization also significantly increases due to more structure factors are integrated, as provided in Tab. \ref{sp}. Among the 20 structure priors based method, the optimal selection achieves the best performance, at the expense of greedy search computation overhead. Our proposed method achieved comparable result but with much less computation burden. Random selection based localization result is the least accurate.
	
	\begin{table}[]
		\centering
		\caption{Structure Priors between Features.}
		\label{sp}
		\begin{tabular}{ |c|l|l| }
			\hline
			& Line                                        & Plane                                                  \\ \hline\hline
			Point & Points on the Line                                 & Points on the Plane                                           \\ \hline
			Line & \begin{tabular}[c]{@{}l@{}}-Parallelism with distance\\ -Orthogonality\end{tabular} & \begin{tabular}[c]{@{}l@{}}-Orthogonality\\ -Parallelsim with distance\\ -Line on the Plane\end{tabular} \\ \hline
			Plane & -                                          & \begin{tabular}[c]{@{}l@{}}-Parallelism with distance\\ -Orthogonality\end{tabular}           \\ \hline
		\end{tabular}
	\end{table}
	\begin{table}[]
		\caption{Average RMSE over 100 Mento-Carlo simulations with different strategies.}
		\centering
		\begin{tabular}{|c|c|c|c|}
			\hline 
			Strategies         & \begin{tabular}[c]{@{}l@{}}Trans.\\ Errors {[}m{]}\end{tabular} & \begin{tabular}[c]{@{}l@{}}Rot. \\ Errors {[}deg.{]}\end{tabular} & \begin{tabular}[c]{@{}l@{}}Time per\\ iteration [s]\end{tabular} \\ \hline\hline
			P-INS          & 0.8186                             & { 6.9399 }                             & { 0.0180 }                         \\ \hline
			PL-INS         & 0.6103                            & 5.0499                             &  0.0193                         \\ \hline
			PLP-INS         & 0.4863                             & 3.9595                             & 0.0274                         \\ \hline
			SPINS-Rand. 20 SPs         & 0.3242                             & 3.1684                             & 0.0259                         \\ \hline
			SPINS-App. Opt. 20 SPs    & 0.2082                             & 3.1652                              & 0.0301                          \\ \hline
			SPINS-Opt. 20 SPs      & 0.1757                             & 2.0399                             & 0.3175                          \\ \hline
			SPINS-All SPs  & 0.1277                             & 1.6984                              & 0.2463                          \\ \hline
			Prior Map & 0.0547  & 0.9414 &--- \\ 
			\hline 
		\end{tabular}
	\end{table}

	\subsection{Euroc dataset}
	In this part, the proposed SPINS framework is tested on the public Euroc MAV Dataset \cite{burri2016euroc}.
	The front-end detection and track point, line and planes based on stereo vision measurements. { Specifically, the point features are detected and tracked with the KLT based optical flow method similar to \cite{Qin2018VINSMono}. The line features are detected and tracked based on a modified line segment detector (LSD) as described in \cite{fu2020pl}. Moreover, the planes are extracted and tracked based on triangulated point features based on the method described in \cite{Nardi2019Unified}.}
	
\begin{figure}
		\centering
		\includegraphics[width=0.75\linewidth]{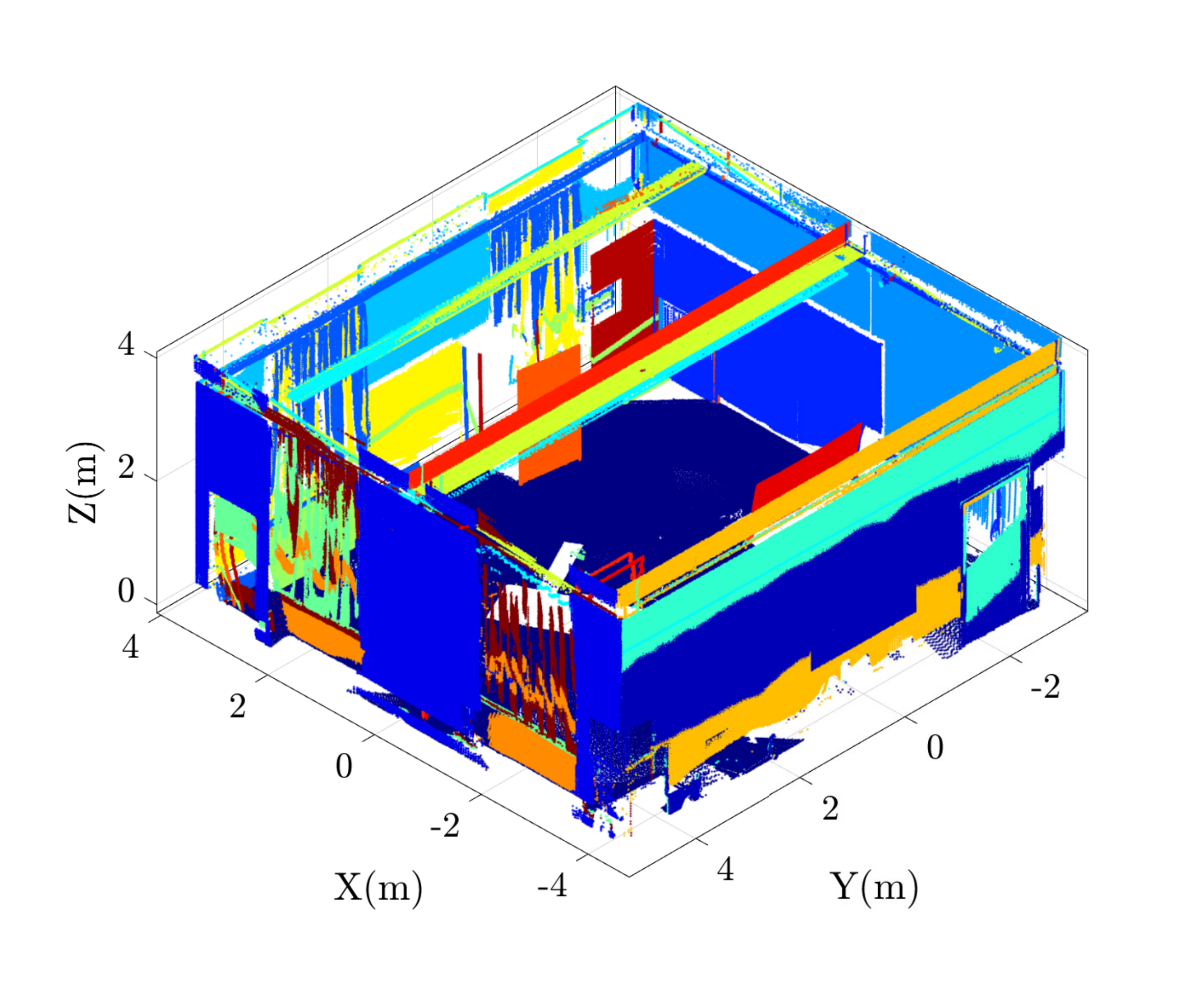}
		\caption{The structure priors extracted from the ground-truth scan of the vicon room.}
		\label{fig:room_planes-crop}
	\end{figure}

\begin{figure}
		\centering
		\includegraphics[width=0.75\linewidth]{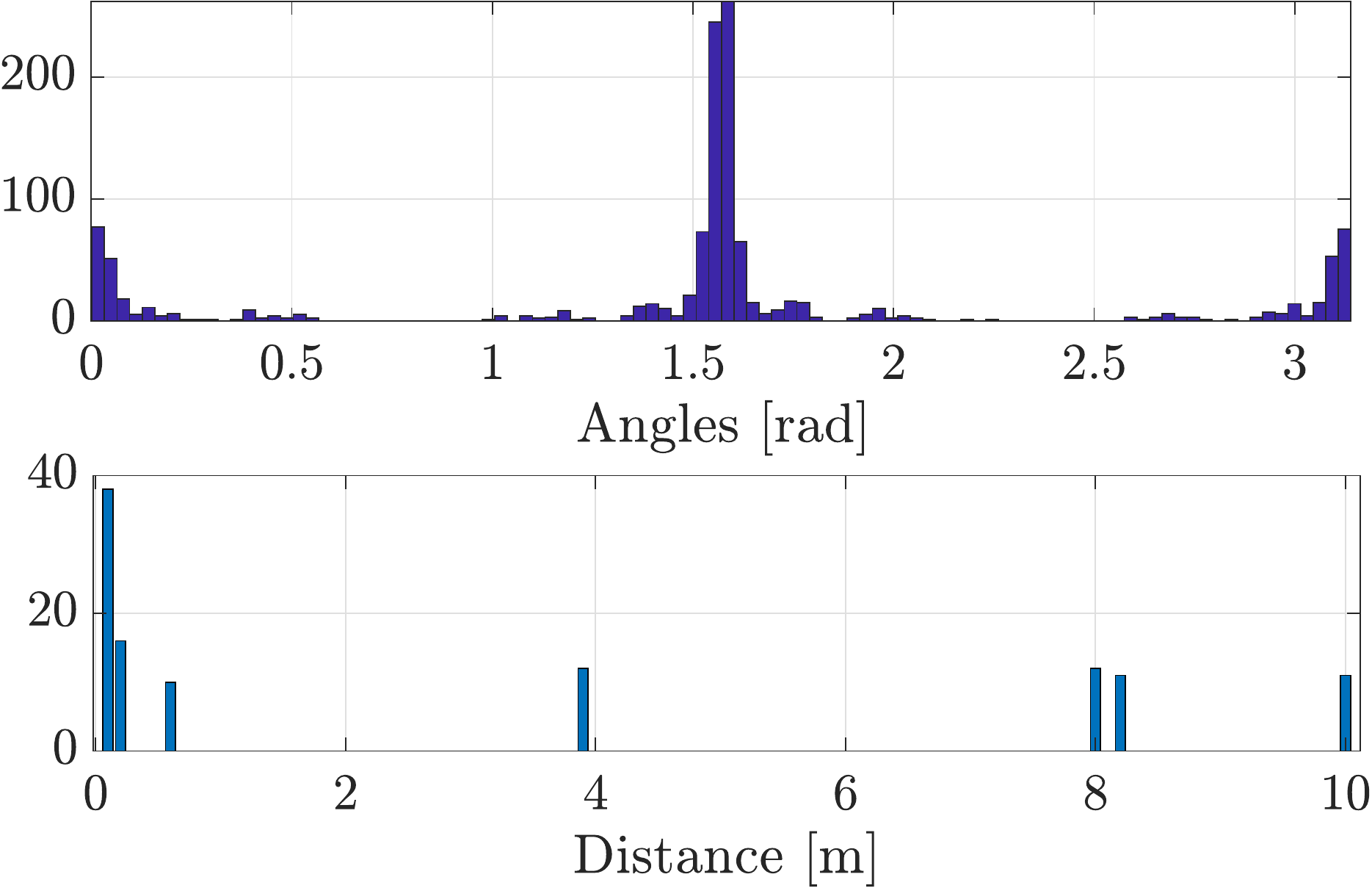}
		\caption{The angles and distances distribution of segemented planes in Fig.\ref{fig:room_planes-crop}}.
		\label{fig:distributions}
	\end{figure}

	On the prior information part, we use the point cloud from the Vicon room to obtain the plane related structure priors. The plane extraction based on the point cloud is shown in Fig. \ref{fig:room_planes-crop}. The corresponding distributions of angles between planes and the distances of parallel planes are plotted in Fig. \ref{fig:distributions}, which show the repetition and sparsity angle and distance pattern in a man-made environment. { Based on the prior information above, we extract the distance priors, such as point-on-plane, line-on-plane, and plane-to-plane distances, and angles priors, such as plane parallel, plane orthogonality, as structure priors to aid the localization.}
	
	The localization results by implementing the VINS FUSION(\url{https://github.com/HKUST-Aerial-Robotics/VINS-Fusion.git}), ORB-SLAM3 (\url{https://github.com/UZ-SLAMLab/ORB_SLAM3.git}), the heterogeneous features based method\cite{yang2019observability}, and the proposed framework are listed in Table \ref{rmse} based on the evaluation method described in \cite{Zhang18iros}. {\color{blue} We use the same experiment setup for all tests according to the Euroc dataset parameters.}
	As indicated, our proposed method can achieve the best performance in both translational and rotational RMSE in $\mathtt{V1\_01}$, $\mathtt{V2\_01}$, $\mathtt{V2\_02}$ and $\mathtt{V2\_03}$. Specifically, our proposed SPINS outperforms the PLP based method\cite{yang2019observability} in all datasets, which shows the effectiveness of incorporating structure prior information. 

\begin{figure}
		\centering
		\includegraphics[width=0.75\linewidth]{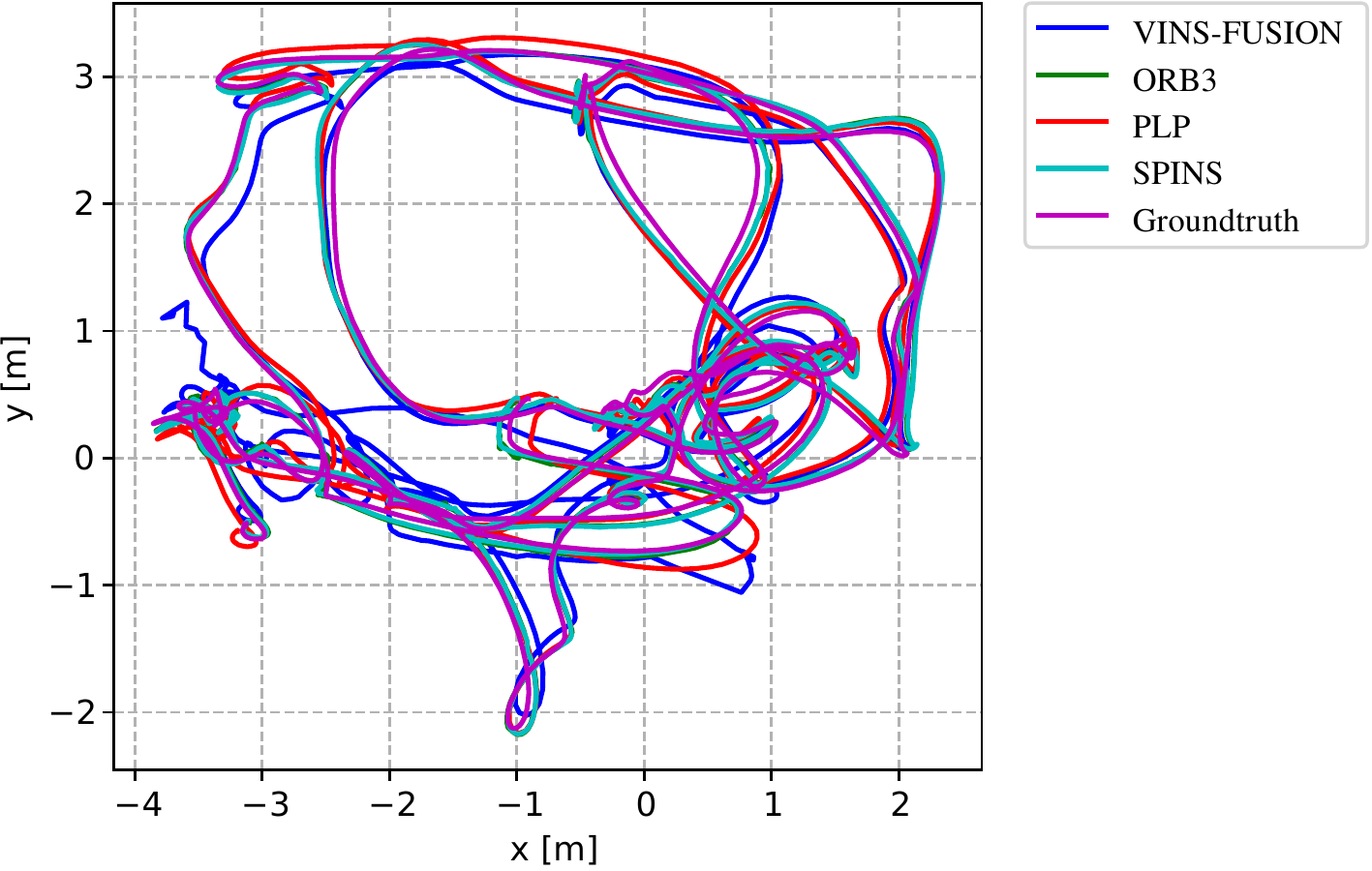}
		\caption{The estimated trajectories of $\mathtt{V1\_03}$ based on the VINS-FUSION, ORB3 and the proposed SPINS with PLP and PLP-SP setups.}
		\label{fig:V103_trajectory_top-crop}
	\end{figure}

	As one example, the results of estimated trajectories of $\mathtt{V2\_02}$ are plotted in Fig. \ref{fig:V103_trajectory_top-crop}. 
	
	\begin{table}[tbh]
		\caption{The RMSE of the estimation results based on different methods}
		\label{rmse}
		\centering
		\begin{tabular}{|c|c|c|c|c|}
			\hline 
			\multirow{2}{*}{Data} & \multicolumn{4}{c|}{Trans. RMSE [m]/Rot. RMSE [$^\circ$]} \\ \cline{2-5} 
			& VINS  & ORB3   & PLP     & SPINS    \\ \hline\hline
			V1\_01          & 0.129/1.748 & 0.085/1.484 & 0.098/1.674 & {\bf 0.079}/{\bf 1.131} \\ \hline
			V1\_02          & 0.145/1.504 & {\bf0.089}/1.336 & 0.321/1.455 & { 0.094}/{\bf 0.905} \\ \hline
			V1\_03          & 0.144/1.967 & {\bf 0.093}/1.952 & 0.193/2.389 & 0.095/{\bf 1.762} \\ \hline
			V2\_01          & 0.150/3.121 & 0.085/1.852 & 0.116/2.352 & {\bf 0.077/1.731} \\ \hline
			V2\_02          & 0.197/4.413 & { 0.167/3.141} & 0.188/4.581 & {\bf 0.160/3.030} \\ \hline
			V2\_03          & 0.219/2.924 & 0.160/3.007 & 0.213/3.458 & {\bf 0.151/2.988} \\ \hline
		\end{tabular}
	\end{table}
	
	\begin{figure*}
		\centering
		\includegraphics[width=1\linewidth]{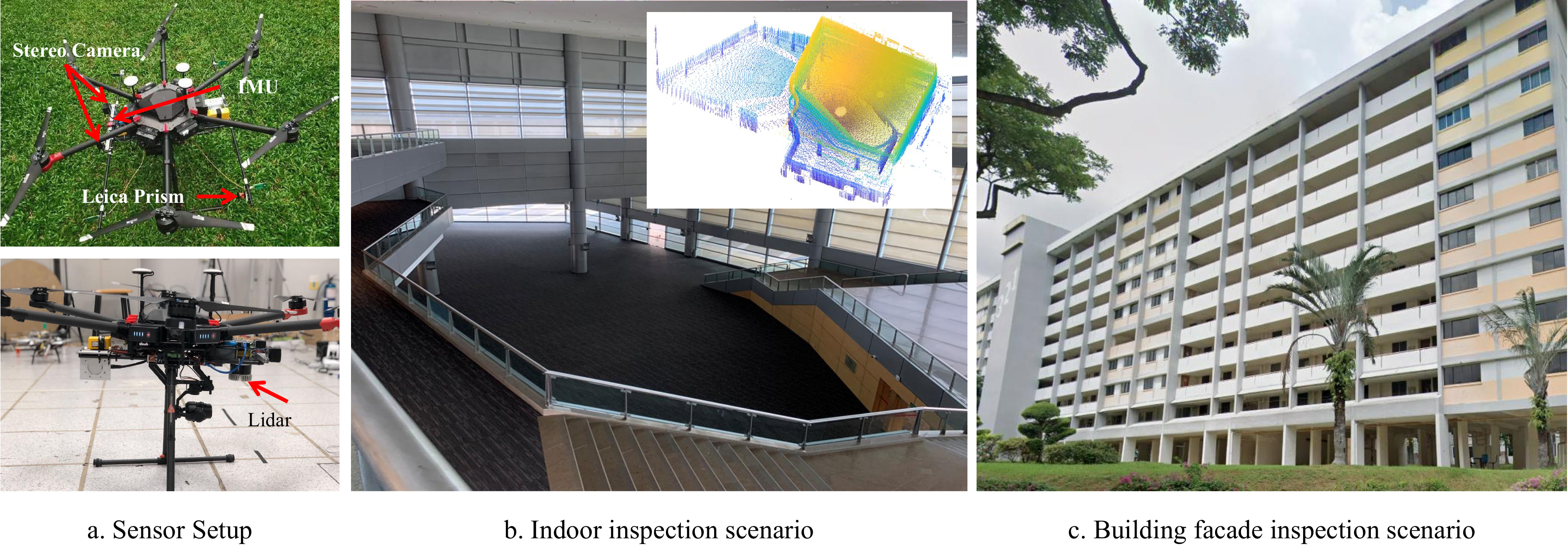}
		\caption{The experiment setups. We test the SPINS in both indoor (b) and outdoor (c) inspection scenarios.}
		\label{fig:expsetup-crop}
	\end{figure*}
	
	\begin{figure}[htb]
		\centering
		\includegraphics[width=0.4\linewidth]{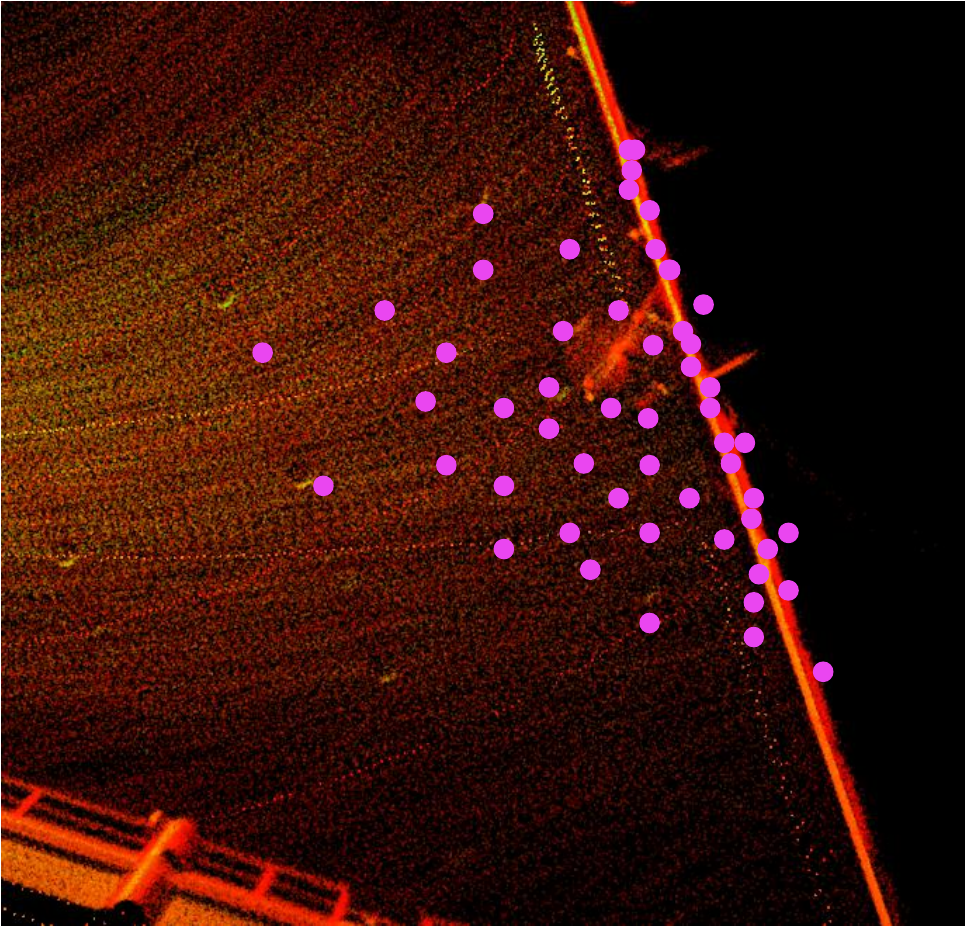}
		\includegraphics[width=0.4\linewidth]{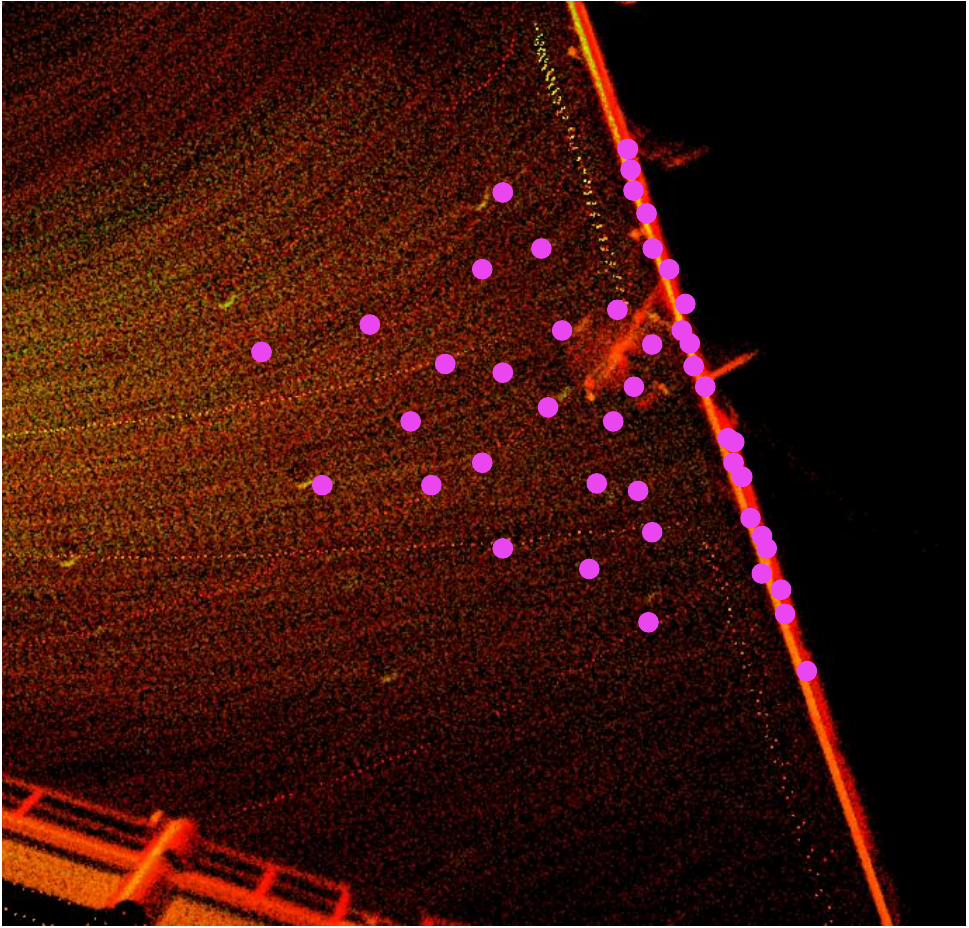}
		\caption{The point feature estimation before/after integrate the point-on-line constraint. When the constraint is imposed, the points on the wall are more accurately stick to the wall, lead to better point feature estimation.}
		\label{fig:after-crop}
	\end{figure}
	\subsection{Field collected data}
	In this part, the proposed SPINS is further tested on large scale inspection environment. A DJI M600 pro hexacopter carrying various sensors is utilized to detect features of the environment, as illustrated in Fig.\ref{fig:expsetup-crop}(a). We considered two scenarios where geometric feature and structure patterns are rich, as shown in Fig.\ref{fig:expsetup-crop}(b) indoor navigation and Fig.\ref{fig:expsetup-crop}(c) building fa\c{c}ade inspection, which is available as part of the VIRAL dataset\cite{nguyen2021ntu}. 
	
	We collect sensing data from visual cameras, LiDARs, and IMU sensors to properly detect the geometric features based on similar front-end processing to Sec. VII.B. {\color{blue} The ground truth is provided by a Leica Geosystem that measures the optical prism onboard. All tests are carried out based on the same parameter setup provided in \url{https://ntu-aris.github.io/ntu_viral_dataset/}.}
	{ \subsubsection{Indoor navigation}
		In this part, the proposed structure prior information is further tested on an indoor auditorium. Similarly, we measure some potentially repetitive and salient features as the structure prior information. Moreover, we use the Leica system to obtain a point cloud map to extract plane based structure priors similar to Sec. VII.B. Three different trajectories are generated to test the proposed methods. 
		
		We compare our results to methods based on monocular camera (VINS-Mono \cite{Qin2018VINSMono}), stereo-camera (VINS-Stereo \cite{qin2019general}), Lidar (LOAM \cite{zhang2014loam}), Lidar and Camera (LVI-SAM \cite{shan2021lvi}), a map based localization method (DLL \cite{caballero2021dll}), and our method (SPINS). The ICP method by registering Lidar scans to point cloud map is also provided as a localization benchmark. More specifically, we use the LOAM odometry for ICP and DLL initialization. The paramters are set as  50 iterations and $0.05$m max correspondence distance in ICP. The estimated trajectories of 3 trials based on the above methods are plotted in Fig.\ref{fig:trial123-crop}. The estimation errors of NYA03 is given in Fig. \ref{fig:errors-crop}. The position RMSE based on the methods are provided in TABLE \ref{rmse_2}. It's apparent, our proposed method, aided by the angle/distance priors (as plotted in Fig. \ref{fig:numsp-crop}), can achieve the best performance among methods mentioned above. Moreover, the distances structure priors are more favored in most cases comparing to angles priors. One example of the effectiveness by imposing point-on-plane constraint, or zero point-to-plane distance constraints, is shown in Fig. \ref{fig:after-crop}.
		
		{Additionally, we compare our method to the prior map based method DLL. Although the DLL method can achieve very close performance to the SPINS in both accuracy and time efficiency, it require an odometery to provide the initialization for registration between map and local scan, which is an extra computation burden. In addition, the DLL method requires an initial position of the robot in the map, which may not be available in practical scenarios. 
			The ICP based method is also provided to indicate the best localization performance that map-based method can achieve. However, the ICP method uses more than 2 seconds for each registration process, and is hard to be implemented in real time applications. }
		\begin{figure*}
			\centering
			\includegraphics[width=1\linewidth]{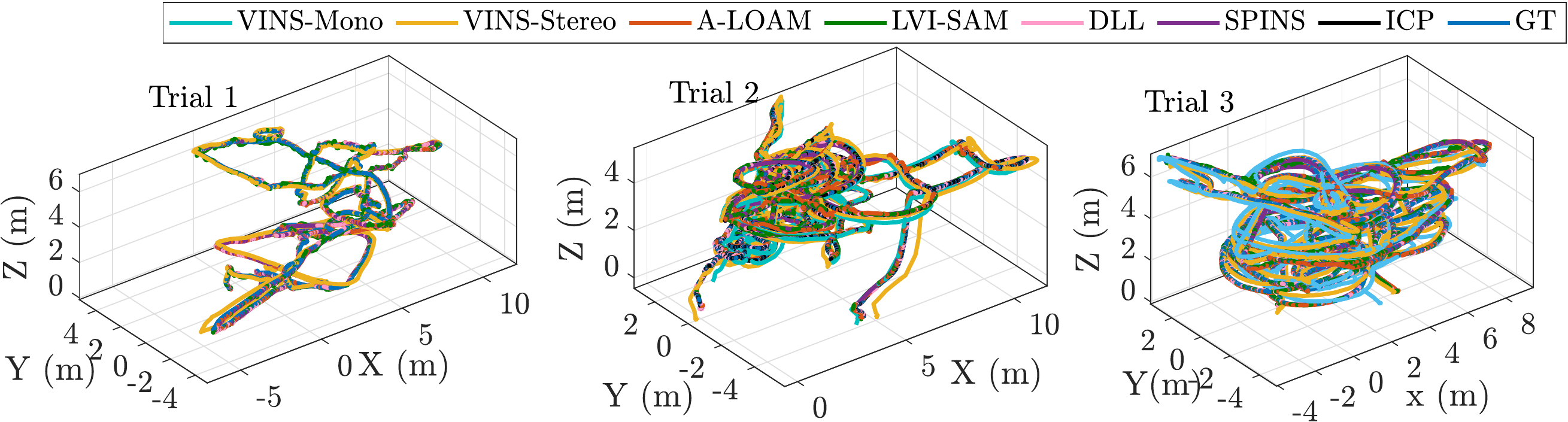}
			\caption{The estimated trajectories based on different methods.}
			\label{fig:trial123-crop}
		\end{figure*}
		\begin{figure}
			\centering
			\includegraphics[width=0.75\linewidth]{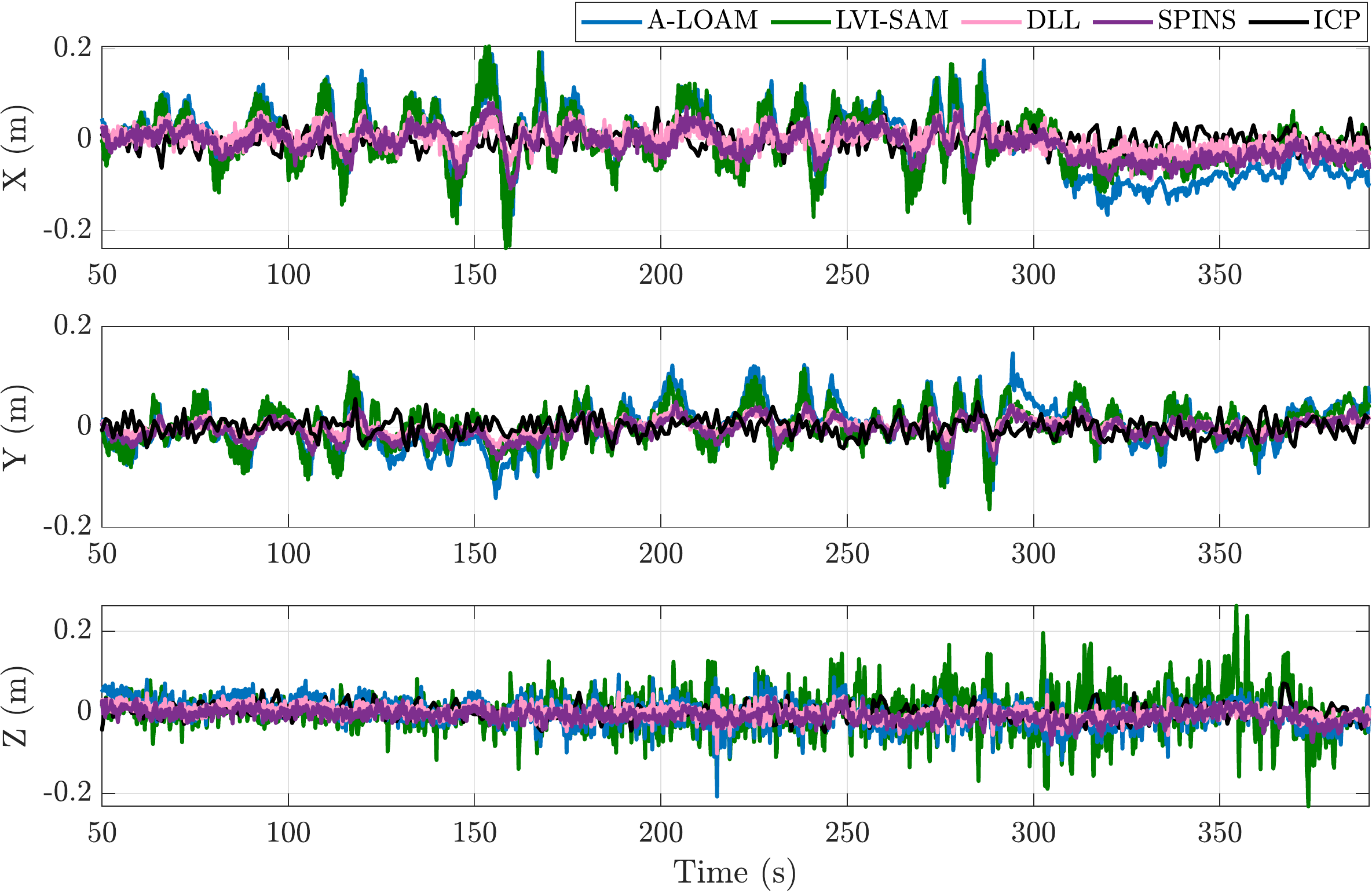}
			\caption{The estimation errors of different methods.}
			\label{fig:errors-crop}
		\end{figure}
		\begin{figure}
			\centering
			\includegraphics[width=0.75\linewidth]{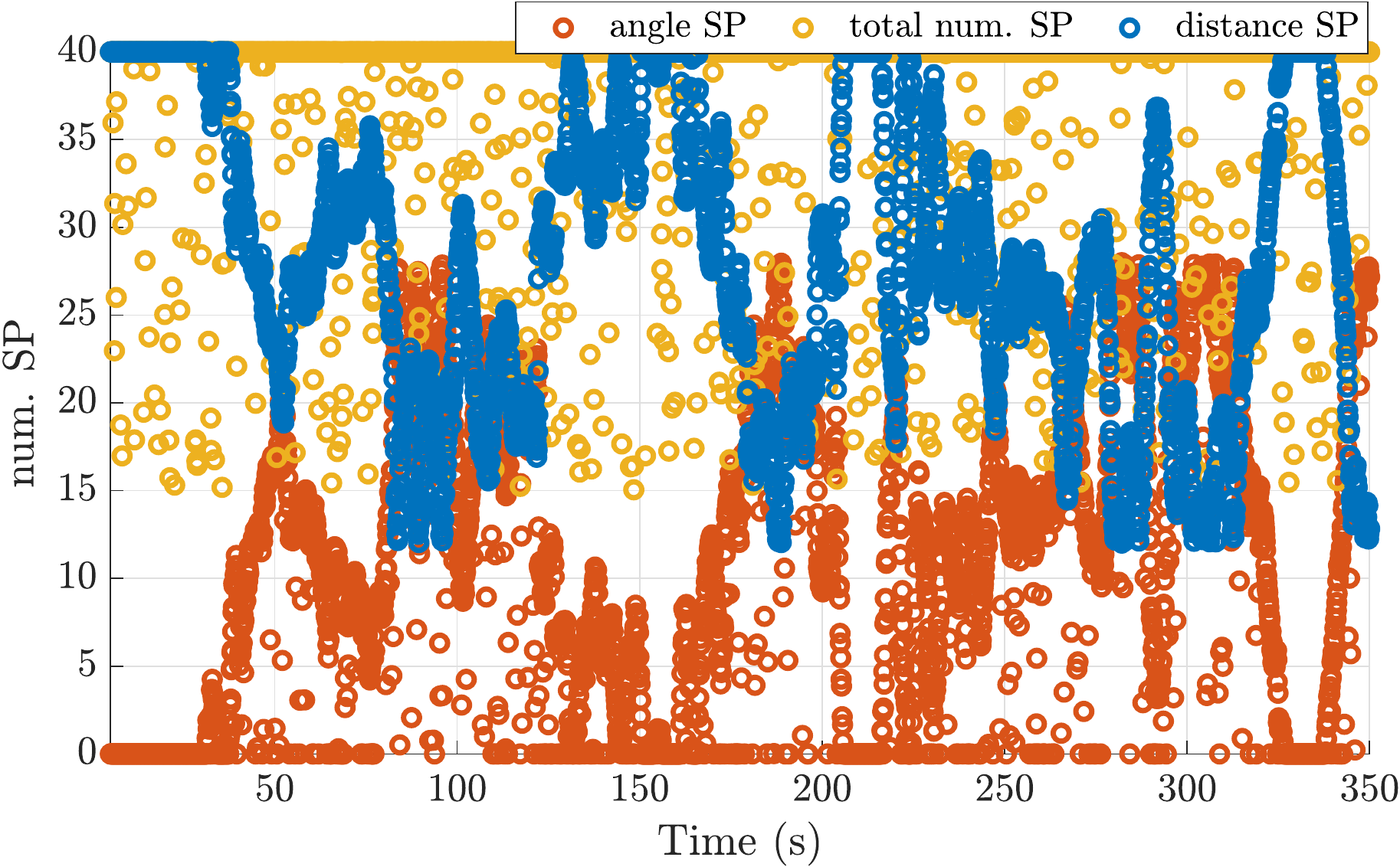}
			\caption{The number of angle and distance structure priors}
			\label{fig:numsp-crop}
		\end{figure}
		\begin{table}[tbh]
			\caption{The RMSE of the estimation results based on different methods}
			\label{rmse_2}
			\centering
			\begin{tabular}{|c|c|c|c|c|}
				\hline
				\multirow{2}{*}{Methods} & \multicolumn{3}{c|}{Translational RMSE {[}m{]}} & \multirow{2}{*}{\begin{tabular}[c]{@{}c@{}}Time per \\ iteration {[}s{]}\end{tabular}} \\ \cline{2-4}
				& NYA01           & NYA02         & NYA03         &                                                                                        \\ \hline\hline
				Vins-Mono                & ------          & 0.2576        & 0.6118        &   ------                                                      \\ \hline
				Vins-Stereo              & 0.2427          & 0.2424        & 0.3808        &   ------                                                      \\ \hline
				ALOAM                    & 0.0768          & 0.0902        & 0.0797        &   ------                                                      \\ \hline
				LVI-SAM                  & 0.0761          & 0.0885        & 0.0827        &   ------                                                      \\ \hline
				DLL                      & 0.0734          & 0.0663        & 0.0607        & 0.0911                                                        \\ \hline
				SPINS                    & 0.0551          & 0.0672        & 0.0592        & 0.0798                                                        \\ \hline
				ICP                      & 0.0262          & 0.0253        & 0.0214        & 2.003                                                         \\ \hline 
			\end{tabular}
			
	\end{table}}
	
	\subsubsection{Building inspection}
\begin{figure}
		\centering
		\includegraphics[width=0.75\linewidth]{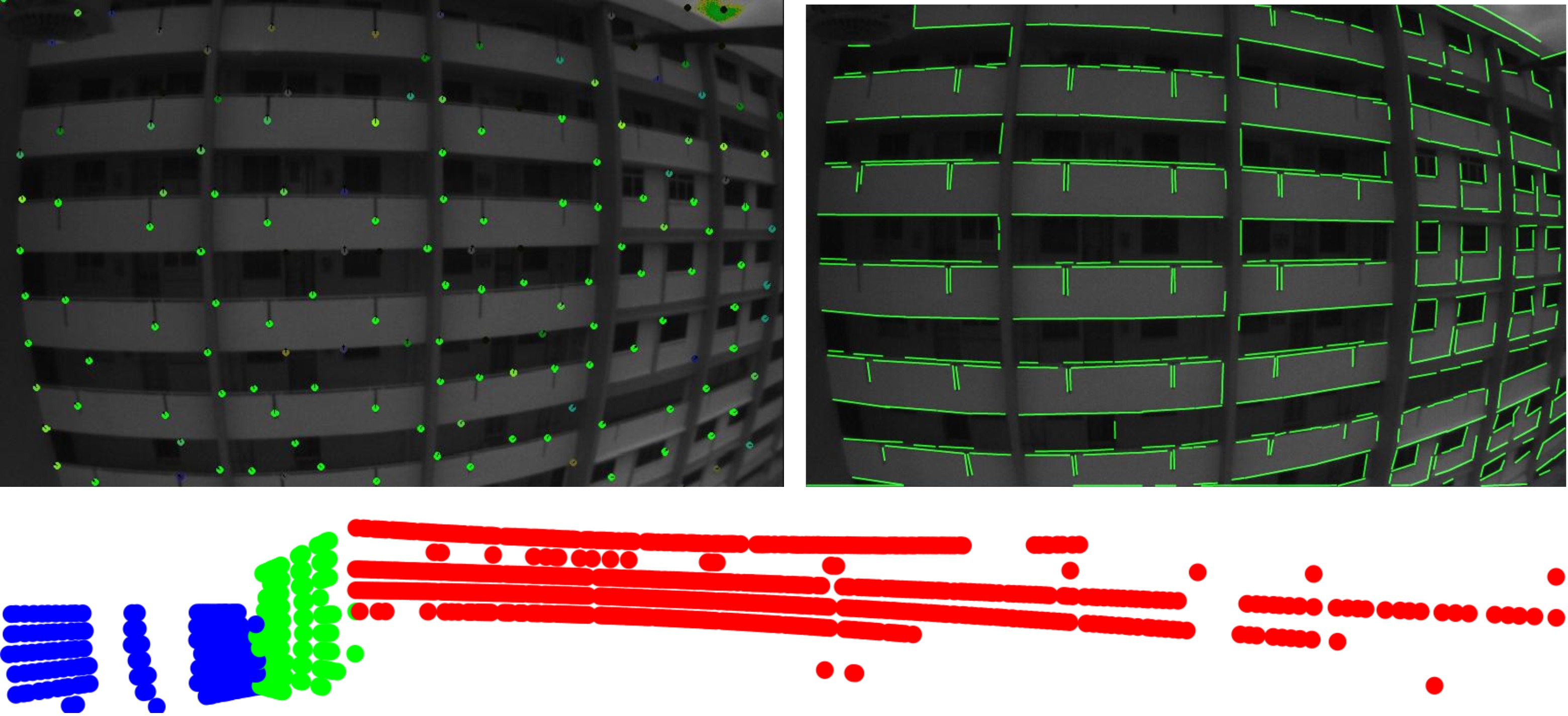}
		\caption{The extraction of point, line and plane features from both vision and LiDAR.}
		\label{fig:sensing}
	\end{figure}

	In the larger scale building inspection task, the UAV is driven to follow a trajectory covering the fa\c{c}ade of the building. The geometric feature extraction is shown in Fig. \ref{fig:sensing}. To utilize our proposed SPINS method, we manually measure some distances and angles, which we treated as main patterns of the building, are summarized as Table \ref{metrics}.
	\begin{table}[]
		\caption{Hand-measured distances for the building}
		\label{metrics}
		\centering
		\begin{tabular}{|l|l|}
			\hline 
			Distance Type& Typical Value {[}m{]}         \\ \hline\hline
			Parallel Lines     & {[}0.3, 0.5, 1.2, 1.5, 2.5, 3.3, 4{]} \\ \hline
			Parallel Line to Plane & {[}0.5, 1.2, 2.7, 3, 4{]}       \\ \hline
			Parallel Planes    & {[}1.5, 3, 4.5, 6{]}         \\ \hline 
		\end{tabular}
	\end{table}
	
\begin{figure}
		\centering
		\includegraphics[width=0.6\linewidth]{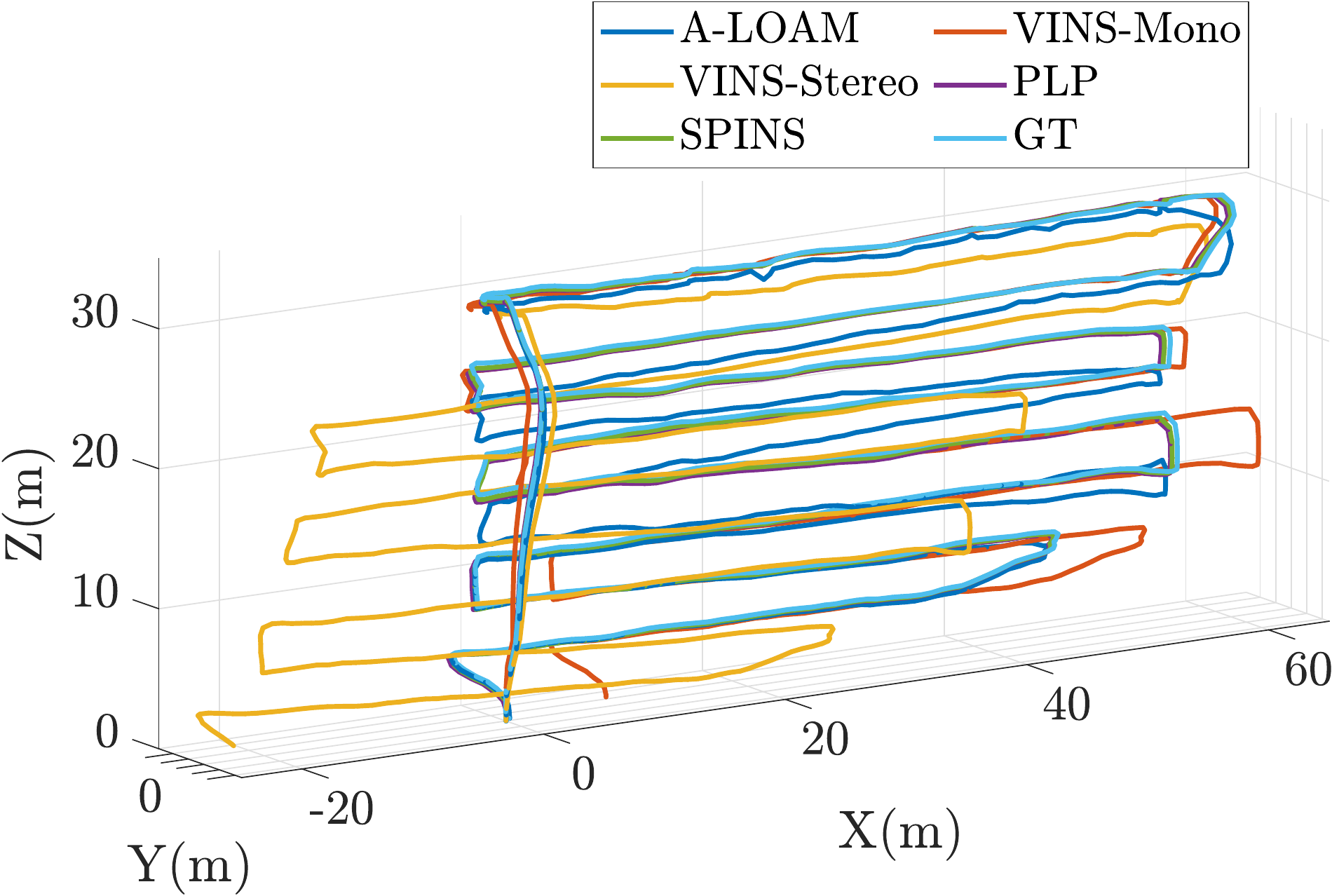}
		\caption{The estimated trajectories from different methods.}
		\label{fig:traj3d-crop}
	\end{figure}

\begin{figure}
		\centering
		\includegraphics[width=0.6\linewidth]{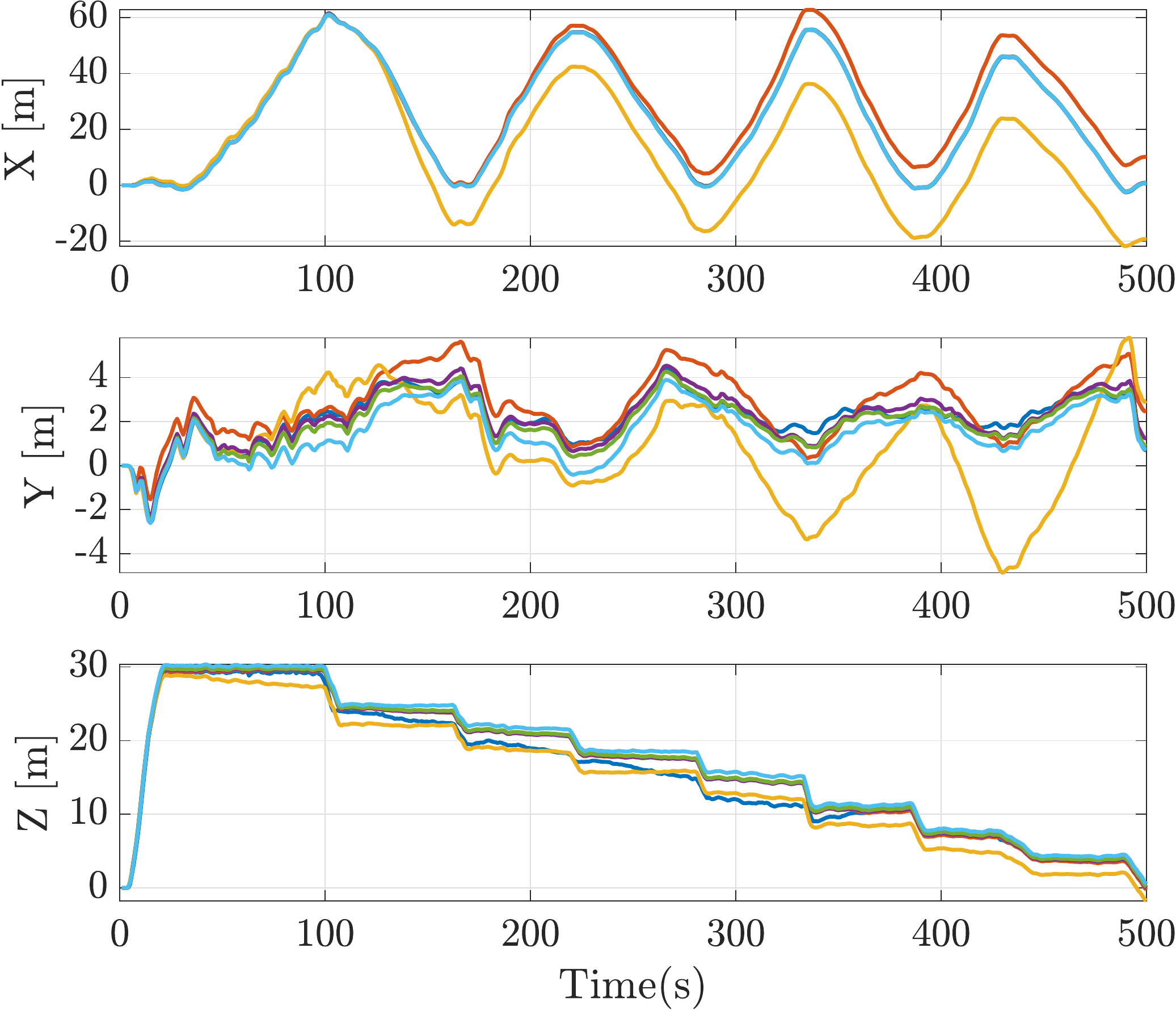}
		\caption{The estimation on X,Y,Z directions.}
		\label{fig:traj-subfigs-crop}
	\end{figure}

\begin{figure}
		\centering
		\includegraphics[width=0.75\linewidth]{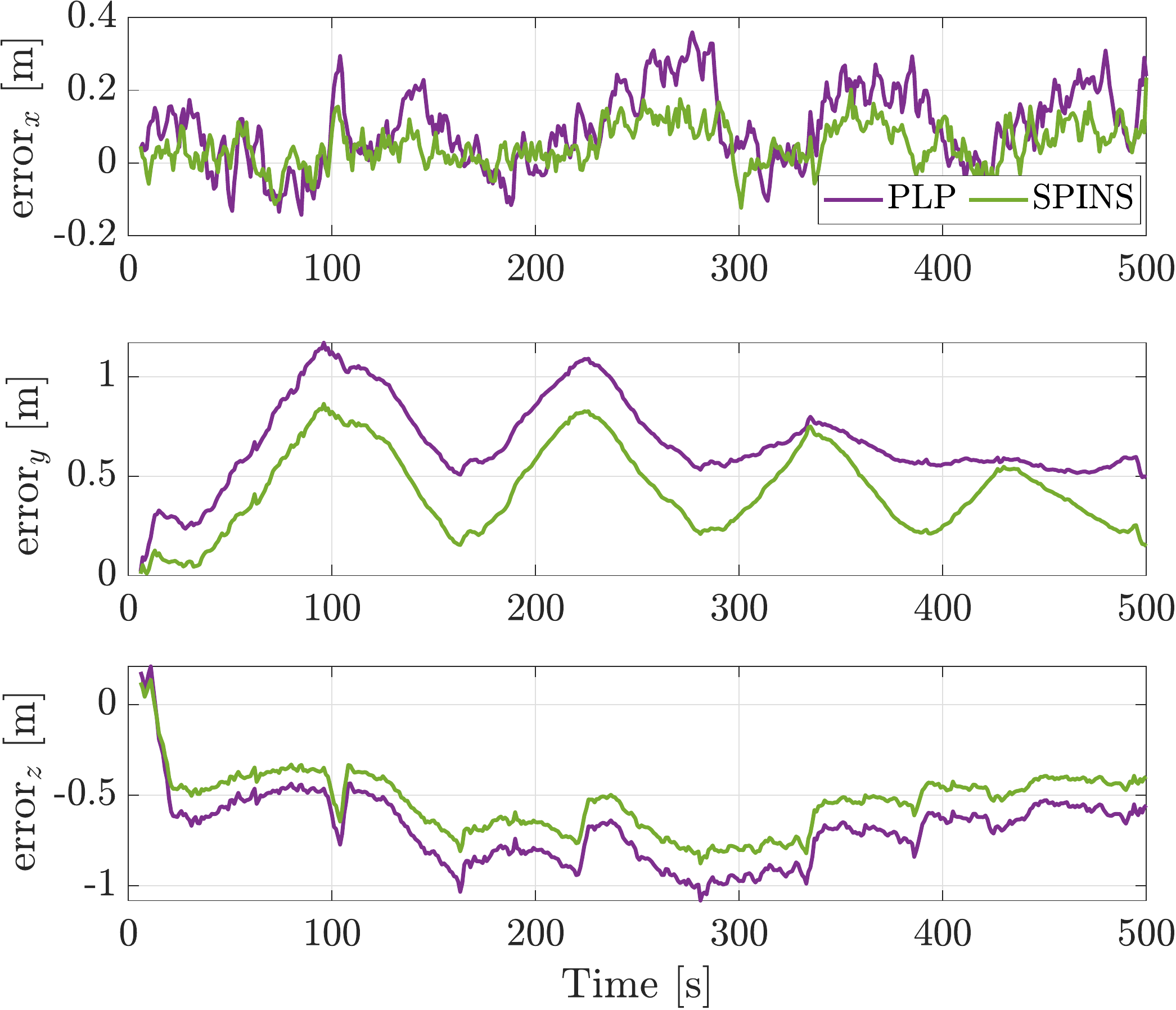}
		\caption{The localization errors on X,Y,Z directions.}
		\label{fig:errors5eps-crop}
	\end{figure}

	The localization results of the inspection using different methods are presented from Fig. \ref{fig:traj3d-crop} to Fig. \ref{fig:errors5eps-crop}. {A demonstrative video} is provided at \url{https://youtu.be/p-wca_WekvQ}. 
	The trajectories of different localization methods along with the groudtruth (GT) are plotted in Fig. \ref{fig:traj3d-crop}, although all trajectories are initialized at the same $[0,0,0]^\top$, the trajectories from the VINS drifts as time goes on. Specifically, from Fig. \ref{fig:traj-subfigs-crop}, it's clear that the trajectories from VINS drifts in $x$ and $y$ directions due to the low feature density and variation during the horizontal movement, and on the other hand, the trajectory from A-LOAM drifts mainly in the $z$ direction due to low depth variation during vertical movement in $z$ direction movement in a 2.5D building. As indicated in Fig. \ref{fig:errors5eps-crop}, the results obtained by using point, line and plane features apparently have better performance in all three directions. Moreover, the integration of structure priors SPINS outperformed the PLP method with the provide structure prior information. The position estimation RMSE of the PLP and SPINS are $1.018$m and $0.7452$m, respectively, with a significant position accuracy improvement by incorporating the structure priors.
	% A demonstrative video can be found in \url{youtubevideolink}.
	
	\section{Conclusions}
	\label{conclusion}
	We propose a sliding window optimization-based localization framework utilizing point, line, and plane features. Considering heterogeneous features, we further develop a structure priors integration method that can further improve localization robustness and accuracy. To alleviate the computation burden brought by extra structure factor edges in the factor graph, we adopt a screening mechanism to select the most informative structure priors. 
	{\color{blue}Although the proposed SPINS shows advantages in civilian environments in our experiment, its effectiveness in more generic environments is questionable.
	In the future, we will further investigate on using general semantic prior information to aid the localization to achieve more wide range of environment adaptation. }
	
	\appendix
	\section{Least-Square solver}
	\label{solver}
	In the non-Euclidean space $\mathcal{X}$, the approximation is achieved by expanding the residual around the origin of a chart computed at current estimation $\hat{\mathcal{X}}$ as 
	\begin{equation}
		\label{expansion}
		\begin{aligned}
			\rbf(\hat{\mathcal{X}}\boxplus \Delta \mathcal{X}) &\approx \rbf(\hat{\mathcal{X}}) + \frac{\partial \rbf(\hat{\mathcal{X}}\boxplus \Delta\mathcal{X})}{\partial \Delta \mathcal{X}}|_{\Delta\mathcal{X}=\mathbf{0}}\Delta \mathcal{X}\\
			&= \rbf(\hat{\mathcal{X}})+\Jbf\Delta\mathcal{X}.
		\end{aligned}
	\end{equation}
	The operator $\boxplus$ applies a perturbation $\Delta \mathcal{X}$ to the manifold space $\mathcal{X}$.  Specifically, for the Euclidean states, the $\boxplus$ operator degenerates to the vector addition operator.
	$\Jbf$ is the sparse Jacobian matrix with only none zero block on the related two states.
	Incorporating the approximation (\ref{expansion}) into (\ref{ful_cost}), we have the following form
	\begin{equation}
		\label{approximation}
		\Delta\mathcal{X}^T \Hbf\Delta\mathcal{X} + 2\bbf^T\Delta\mathcal{X}^T + \mathrm{const}(\hat{\mathcal{X}}),    
	\end{equation}
	where 
	\begin{equation*}
		\begin{aligned}
			\Hbf =&\Jbf_p^T\Pbf^{-1}_p\Jbf_p +  \sum_{m\in T_t} \Jbf_{m}^T\Sigma^{-1}_m\Jbf_m\\
			&+ \sum_{i\in P_t} \Jbf_{i}^T\Sigma^{-1}_i\Jbf_i+\sum_{j\in L_t} \Jbf_{j}^T\Sigma^{-1}_j\Jbf_j+\sum_{k\in \Pi_t} \Jbf_{k}^T\Sigma^{-1}_k\Jbf_k\\
			&+\sum_{s\in S_t} \Jbf_{s}^T\Sigma^{-1}_s\Jbf_s,
		\end{aligned}
	\end{equation*}
	and 
	\begin{equation*}
		\begin{aligned}
			\bbf =& \rbf_p^T\Pbf_p^{-1}\Jbf_p^T + \sum_{m\in T_t} \rbf_{m}^T\Sigma^{-1}_m\Jbf_m\\
			&+ \sum_{i\in P_t} \rbf_{i}^T\Sigma^{-1}_i\Jbf_i+\sum_{j\in L_t} \rbf_{j}^T\Sigma^{-1}_j\Jbf_j+\sum_{k\in \Pi_t} \rbf_{k}^T\Sigma^{-1}_k\Jbf_k\\
			&+\sum_{s\in S_t} \rbf_{s}^T\Sigma^{-1}_s\Jbf_s,
		\end{aligned}
	\end{equation*}
	and $\text{const}(\hat{\mathcal{X}})$ is a constant term depending only on $\hat{\mathcal{X}}$.
	
	As a result, (\ref{approximation}) can be minimized by solving 
	\begin{equation}
		(\Hbf + \lambda\Ibf)\Delta \mathcal{X} = -\bbf, 
	\end{equation} 
	where $\lambda$ is a damping factor.
	The estimation is hereafter updated as 
	\begin{equation}
		\hat{\mathcal{X}} = \hat{\mathcal{X}} \boxplus \Delta{\mathcal{X}}.
	\end{equation}
	The update iterates until convergence.
	
	\section{Jacobians of Heterogeneous Feature Measurements}
	\subsection{Point Measurement Jacobians}
	\label{point_feature}
	The Jacobians, based on the formulations in (\ref{expansion}), are related to the state of the point in Euclidean space $\xbf_i = {^G}\pbf_i$, and the pose of the robot in $\mathtt{SE}(3)$, and can be calculated as follows.
	\begin{equation}
		\Jbf_{\hat\xbf_i}^{\rbf_i} = \left.\frac{\partial \rbf_i(\hat\xbf_i + \Delta \xbf_i)}{\partial \Delta \xbf_i}\right|_{\Delta \xbf_i= \mathbf{0}} = {^I_G}\hat{\Rbf},
	\end{equation}
	and 
	\begin{equation}
		\begin{aligned}
			\Jbf_{\hat \pbf_I}^{\rbf_i} &= \left.\frac{\partial \rbf_i(\hat\xbf_I \boxplus \Delta \xbf_I)}{\partial \delta \pbf_I}\right|_{\Delta \xbf_I= \mathbf{0}} = -	{^I_G}\hat\Rbf,\\
			\Jbf_{\hat{\bar{q}}_I}^{\rbf_i} &= \left.\frac{\partial \rbf_i(\hat\xbf_I \boxplus \Delta \xbf_I)}{\partial \delta\pmb\theta_I}\right|_{\Delta \xbf_I= \mathbf{0}} = - {^I_G}\hat\Rbf [\hat\pbf_i ]_{\times},
		\end{aligned}
	\end{equation}
	where $\Delta\xbf_i$ and $\Delta \xbf_I$ are the perturbations on $\hat \xbf_i$ and $\hat \xbf_I$, respectively.  Specifically, $\begin{bmatrix}
		\delta\pmb\theta_I^\top, \delta\pbf_I^\top
	\end{bmatrix}^\top$ are the rotational and translational part of the perturbation $\Delta \xbf_I$. The measurement Jacobians over other parts of $ \xbf_I$ are $\mathbf{0}$.
	\subsection{Line Measurement Jacobians}
	\label{line_feature}
	The measurement residual Jacobian w.r.t the local pose is 
	\begin{equation}\begin{aligned}
			\Jbf^{\rbf_j}_{\hat{\bar{q}}_I} &=	\left.\frac{\partial {\mathbf{r}}_j(\hat\xbf_I \boxplus \Delta \xbf_I)}{\partial \delta\pmb\theta_{I}}\right|_{\Delta \xbf_I = \mathbf{0}}\\ &=\left[\begin{array}{cc}
				\left[_{G}^{I} \hat{\mathbf{R}}^{G} \hat{\mathbf{n}}_{j}\right]_{\times}-\left[_{G}^{I} \hat{\mathbf{R}}\left[^{G} \hat{\mathbf{p}}_{I}\right]_{\times} {^G}\hat{\mathbf{v}}_{j}\right]_{\times} \\
				-\left[_{G}^{I} \mathbf{\hat { R }}^{G} \hat{\mathbf{v}}_j\right]_{\times}
			\end{array}\right], \\
			\Jbf^{\rbf_j}_{\hat{\pbf}_I} &\doteq\left.	\frac{\partial\rbf_j(\hat\xbf_I \boxplus \Delta \xbf_I)}{\partial  \delta\pbf_{I}}\right|_{\Delta \xbf_I = \mathbf{0}} \\&=\left[\begin{array}{c}
				{_G^I}\hat{\mathbf{R}}\left[^{G} \hat{\mathbf{v}}_{j}\right]_{\times} \\
				\mathbf{0}_{3}
			\end{array}\right].
		\end{aligned}
	\end{equation}
	The measurement residual Jacobians w.r.t. to line state can be calculated using the chain rule as 
	\begin{equation}
		\Jbf_{\hat\xbf_j}^{\rbf_j} = \Jbf_{\hat{\bar{\lbf}}_j}^{\rbf_j}\Jbf_{ \hat\xbf_j}^{\hat{\bar{\lbf}}_j},
	\end{equation}
	where $\bar\lbf_j =\begin{bmatrix}
		\bar{q}_k^\top & d_j
	\end{bmatrix}^\top$ is an intermediate state. Specifically, with perturbation $\delta \bar{\Ibf}_j  = \begin{bmatrix}
		\delta\pmb\theta_j^\top& \delta d_j
	\end{bmatrix}^\top$ on manifold $\bar{\lbf}_j$, we have
	\begin{equation}
		\small
		\begin{aligned}
			\Jbf_{\hat{\bar{\lbf}}_j}^{\rbf_j} &= \left.\frac{\partial\rbf_j(\hat{\bar\lbf}_j\boxplus \delta \bar\lbf_j)}{\partial\delta \bar\lbf_j}\right|_{\delta \bar\lbf_j= \mathbf{0}}\\ &= \left[\begin{array}{cc}
				_{G}^{I}\Rbf & -_{G}^{I} \mathbf{R}\left[^{G} \pbf_{I} \right]_{\times} \\
				\mathbf{0}_{3} & {_{G}^{I}}\mathbf{R}
			\end{array}\right]\left[\begin{array}{cc}
				^G\hat{d}_{j}\left[^{G} \hat{\mathbf{R}}_j \mathbf{e}_{1}\right]_{\times} & ^{G} \hat{\mathbf{R}}_{j} \mathbf{e}_{1} \\
				\left[^{G} \hat{\mathbf{R}}_{j} \mathbf{e}_{2} \right]_{\times} & \mathbf{0}_{3 \times 1}
			\end{array}\right],
		\end{aligned}
	\end{equation}
	and   
	\begin{small}
		\begin{equation}
			\label{cpdev}
			\Jbf_{\hat{\xbf}_j}^{\hat{\bar{\lbf}}_j} = \left.\frac{\partial \bar\lbf_j(\hat \xbf_j+\Delta\xbf_j)}{\partial \Delta\xbf_j}\right|_{\Delta \xbf_j = \mathbf{0}}=\left[\begin{array}{cc}
				\frac{2}{\hat{d}}\left(\hat{q}_{j} \mathbf{I}_{3}-\left[\hat{\mathbf{q}}_{j}\right]_{\times}\right) & -\frac{2}{\hat{q}} \hat{\mathbf{q}}_{j} \\
				\hat{\mathbf{q}}_{j}^{\top} & \hat{q}_{j}
			\end{array}\right],
		\end{equation}
	\end{small}
	\noindent
	where $\ebf_1, \ebf_2$ are the first and second column of the identity matrix $\Ibf_3$. $\qbf_j$ and $q_j$ are the vector part and scalar part of the quaternion $\bar{q}_j$, respectively. Please refer to \cite{yang2019aided} for detailed derivation of (\ref{cpdev}).
	\subsection{Plane Measurement Jacobians}
	\label{plane_feature}
	By defining an intermediate state of the plane as $\bar\pbf_j = \begin{bmatrix}
		{^G}\nbf_k^\top & {^G}d_k
	\end{bmatrix}^\top$, 
	the measurement residual $\rbf_k$'s Jacobian w.r.t. the plane state $\xbf_k = {^G}\pbf_k$ can be calculated using the chain rule as
	\begin{equation}
		\label{jacobian_plane}
		\Jbf^{\rbf_k}_{\hat\xbf_k}= \Jbf^{\rbf_k}_{\hat{\bar{\pbf}}_k}\Jbf^{\hat{\bar\pbf}_k}_{\hat\xbf_k},
	\end{equation}
	where 
	\begin{small}
		\begin{equation}
			\label{interjacb}
			\begin{aligned}
				\Jbf^{\rbf_k}_{\hat{\bar{\pbf}}_k} &= \left.\frac{\partial \rbf_k(\hat{\bar\pbf}_k\boxplus\delta\bar\pbf_k)}{\partial \delta\bar\pbf_k}\right|_{\delta\bar\pbf_k=\mathbf{0}} \\&= \begin{bmatrix}
					{_{G}^{I}}\hat{\mathbf{R}}\left(\left(^G \hat{d}_{k}-{^G} \hat{\mathbf{n}}_{k}^{\top G} \hat{\mathbf{p}}_{I}\right) \mathbf{I}_{3}- {^G}\hat{\mathbf{n}}_{k} 
					{^G}\hat{\mathbf{p}}_{I}^{\top}\right) & {_{G}^{I}}\hat{\mathbf{R}}^{G} \hat{\mathbf{n}}_{k}
				\end{bmatrix},\\
				\Jbf^{\hat{\bar\pbf}_k}_{\hat\xbf_k} &=\left.\frac{\partial \bar\pbf_k(\hat\xbf_k+\Delta\xbf_k)}{\partial \Delta\xbf_k}\right|_{\Delta \xbf_k = \mathbf{0}} = \begin{bmatrix}
					\frac{\left(
						\mathbf{I}_{3}-^{G} \hat{\mathbf{n}}_{k}^{G} \hat{\mathbf{n}}_{k}^{\top}\right)}{^G\hat{d}_{k}} \\
					{^{G}}\hat{\mathbf{n}}_{k}^\top
				\end{bmatrix}.
			\end{aligned}
		\end{equation}
	\end{small}
	$\delta\bar \pbf_k$ and $\Delta \xbf_k$ are perturbations on $\hat{\bar{\pbf}}_k$ and $\hat \xbf_k$, respectively. 
	
	By injecting a small perturbation of local pose into (\ref{plane_GtoI}), we have the measurement Jaocobian w.r.t. the pose as follows:
	\begin{equation}
		\begin{aligned}
			\Jbf^{\rbf_k}_{\hat{\bar{q}}_I} &=\left.\frac{\partial {\rbf_k(\hat\xbf_I \boxplus \Delta\xbf_I)}}{\partial \delta \boldsymbol{\theta}}\right|_{\Delta \xbf_I = \mathbf{0}} \\&=\left(^{G} \hat{d}-{^G} \hat{\mathbf{n}}_{\pi}^{\top G} \hat{\mathbf{P}}_{I}\right)\left[ _{G}^{I} \hat{\mathbf{R}}^{G} \hat{\mathbf{n}}_{\pi} \right]_{\times}, \\
		\end{aligned}
	\end{equation}
	\begin{equation}
		\Jbf^{\rbf_k}_{\hat\pbf_I}= \left.\frac{\partial {\rbf_k(\hat\xbf_I \boxplus \Delta\xbf_I)}}{\partial \delta \pbf_I}\right|_{\Delta \xbf_I = \mathbf{0}} =-_{G}^{I} \hat{\mathbf{R}}^{G} \hat{\mathbf{n}}_{\pi}^{G} \hat{\mathbf{n}}_{\pi}^{\top}.
	\end{equation}
	\section{Jacobians of the Structural Priors Measurements}
	\subsection{Point-Point Measurement Jacobians}
	\label{ptpj}
	For a pair of points $(i,i')\in \mathcal{S}$, the Jacobians of the measurement residual, defined as $\rbf = \zbf- h_{ii'}(\hat{\xbf}_i, \hat{\xbf}_{i'})$, are derived as 
	\begin{align}
		\Jbf^{\rbf_{ii'}}_{\hat\xbf_i} = \Jbf^{\rbf_{ii'}}_{\hat\xbf_{ii'}}\Jbf^{\hat\xbf_{ii'}}_{\hat\xbf_{i}},\\
		\Jbf^{\rbf_{ii'}}_{\hat\xbf_i'} = \Jbf^{\rbf_{ii'}}_{\hat\xbf_{ii'}}\Jbf^{\hat\xbf_{ii'}}_{\hat\xbf_{i'}},
	\end{align}
	where
	$\Jbf^{\hat\xbf_{ii'}}_{\hat\xbf_{i}} = -\Ibf_3$, $\Jbf^{\rbf_{ii'}}_{\hat\xbf_{ii'}}\Jbf^{\hat\xbf_{ii'}}_{\hat\xbf_{i'}} = \Ibf_3$.
	
	Specifically, if we consider $h_{ii'}\left(\cdot\right)$ as a distance measurment function, the Jacobians can be calculated as 
	\begin{align}
		\Jbf^{\rbf_{ii'}}_{\xbf_{ii'}} =\frac{\hat\xbf_{ii'}}{\|\hat\xbf_{ii'}\|}.
	\end{align}

	\subsection{Point-Line Measurement Jacobians}
	\label{ptlj}
	The measurement residual Jacobian with respect to the point $i$ is 
	\begin{equation}
		\Jbf^{\rbf_{ij}}_{
			\hat\xbf_i}  = \Jbf^{\rbf_{ij}}_{\hat\pbf_{ij}}\Jbf^{\pbf_{ij}}_{\hat\xbf_{i}},
	\end{equation}	where $ \Jbf^{\hat\pbf_{ij}}_{\hat\xbf_{i}}  = \begin{bmatrix}
		\hat{\bar\nbf}_j^\top \\ \hat{\bar\nbf}_j^\top\times\hat{\bar\vbf}_j^\top 
	\end{bmatrix}$.
	
	The measurement residual Jacobian with respect to the line estimation error is 
	\begin{equation}
		\begin{aligned}
			\Jbf^{\rbf_{ij}}_{\pbf_j} = \Jbf^{\rbf_{ij}}_{\pbf_{ij}}\Jbf^{\pbf_{ij}}_{\bar\lbf_{j}}\Jbf^{\bar\lbf_{j}}_{\pbf_{j}}, 
		\end{aligned}
	\end{equation}
	where \[\Jbf^{\pbf_{ij}}_{\bar\lbf_{j}}= 
	\begin{bmatrix}
		{^G}\hat\pbf_i[\hat\Rbf_j\ebf_1]_{\times} & {^G}\hat\pbf_i\hat\Rbf_j\ebf_1\\
		{^G}\hat\pbf_i[\hat\Rbf_j \ebf_3]_{\times} & 1
	\end{bmatrix}
	,\] and
	$\Jbf^{\bar\lbf_{j}}_{\pbf_{j}} $ is calculated in (\ref{cpdev}).
	When considering the measurement as point-line distance, $\Jbf^{\rbf_{ij}}_{\pbf_{ij}}= \frac{\hat\pbf_{ij}}{\|\hat\pbf_{ij}\|}.$
	\subsection{Point-Plane Measurement Jacobians}
	\label{ptplj}
	The measurement residual Jacobian with respect to the point estimate error is 
	\begin{equation}\Jbf^{\rbf_{ij}}_{\xbf_i}= 
		{^G}\hat\nbf_k^\top.
	\end{equation}
	
	The measurement residual Jacobian with respect to the plane estimate error is 
	\begin{equation}
		\Jbf^{\rbf_{ij}}_{\hat\xbf_k} = \Jbf^{\rbf_{ij}}_{\hat{\bar\pbf}_k}\Jbf^{\hat{\bar\pbf}_{ij}}_{\hat\xbf_k}, 
	\end{equation}
	where
	\[ \Jbf^{\rbf_{ij}}_{\bar\pbf_k}= \begin{bmatrix}
		^G\hat\pbf^\top_i &1
	\end{bmatrix}^\top,\]
	and $\Jbf^{\hat{\bar\pbf}_{ij}}_{\hat\xbf_k}$ is provided in (\ref{interjacb}).

	\subsection{Line-Line Measurement Jacobians}
	\label{ltlj}
	The Jacobians of the measurement error $\alpha_{jj'}$ and $d_{jj'}$ w.r.t. the estimation error are 
	\begin{equation}
		\begin{aligned}
			\Jbf_{\hat\xbf_j}^{\rbf_{jj'}^\alpha}= \Jbf_{\hat{\bar\lbf}_j}^{\rbf_{jj'}^\alpha}\Jbf^{\hat{\bar\lbf}_j}_{\hat\xbf_j},
		\end{aligned}
	\end{equation} 
	where \[\Jbf_{\bar\lbf_j}^{\rbf_{jj'}^\alpha} = \hat{\bar\vbf}_{j'}^\top\begin{bmatrix}
		[\hat\Rbf_j\ebf_2]_{\times} &\mathbf{0}_{3\times 1}
	\end{bmatrix}.\]
	The measurement residual Jacobian w.r.t. $j'$ can be calculated similarly.
	Specifically, when two lines are parallel, namely $\bar\vbf_j = \pm\bar\vbf_j'$, the distance measurement Jacobian can be calculated as
	\begin{equation}
		\Jbf_{\hat\xbf_j}^{\rbf_{jj'}^d}= \Jbf_{\hat{\bar\lbf}_j}^{\rbf_{jj'}^d}\Jbf^{\hat{\bar\lbf}_j}_{\hat\xbf_j},
	\end{equation}
	where 
	\begin{equation}
		\begin{aligned}
			\Jbf_{\bar\lbf_j}^{\rbf_{jj'}^d}
			= \hat{\bar\vbf}_j'\times \begin{bmatrix}
				d_j	[ \hat\Rbf_j\ebf_1]_{\times}& \hat\Rbf_j\ebf_1
			\end{bmatrix}.
		\end{aligned}
	\end{equation}
	
	\subsection{Line-Plane Measurement Jacobians}
	\label{ltpj}
	The directional correlation measurement Jacobian is as follows
	\begin{equation}
		\Jbf_{\hat\xbf_j}^{\rbf_{jk}^\alpha}=\Jbf_{\hat\Ibf_j}^{\rbf_{jk}^\alpha}\Jbf^{\hat\Ibf_j}_{\xbf_{j}},
	\end{equation} where

	\begin{equation}
		\Jbf_{\hat\Ibf_j}^{\rbf_{jk}^\alpha}=\begin{bmatrix}
			[\hat\Rbf_j\ebf_3]_{\times} &\mathbf{0}_{3\times 1} 
		\end{bmatrix}, 
	\end{equation}and
	\begin{align}
		\Jbf_{\hat\xbf_k}^{\rbf_{jk}^\alpha} = 
		\frac{\hat{\bar\vbf}_j^\top\left(
			\mathbf{I}_{3}-^{G} \hat{\mathbf{n}}_{k}^{G} \hat{\mathbf{n}}_{k}^{\top}\right)}{^G\hat{d}_{k}}.
	\end{align}
	
	When the line $j$ is parallel to the plane, namely $\bar\vbf_j^\top\nbf_k=0$, the distance residual Jacobians 
	\begin{equation}
		\Jbf_{\hat\xbf_j}^{\rbf_{jk}^d}=\Jbf_{\hat\Ibf_j}^{\rbf_{jk}^d}\Jbf^{\hat\xbf_j}_{\xbf_{j}},
	\end{equation}
	where
	\begin{equation}
		\Jbf_{\hat\Ibf_j}^{\rbf_{jk}^d} = \hat{\nbf}_k^\top \begin{bmatrix}
			[\hat \Rbf_j\ebf_2]_{\times} & \hat{\Rbf}_j\ebf_2
		\end{bmatrix},
	\end{equation} and 
	\begin{equation} 
		\Jbf_{\hat\xbf_k}^{\rbf_{jk}^d}=
		% \frac{\tilde d_{jk}}{\tilde \pbf_{k}}= 
		\frac{(\hat{\bar\nbf}_j\times \hat{\bar\vbf}_j)^\top\left(
			%		{^G}\hat{\mathbf{n}}_{\pi}^{\top G} \hat{\mathbf{n}}_{\pi}
			\mathbf{I}_{3}-^{G} \hat{\mathbf{n}}_{k}^{G} \hat{\mathbf{n}}_{k}^{\top}\right){^G}\hat d_j}{^G\hat{d}_{k}} + {^G}\hat\nbf_k^T.
	\end{equation}
	\subsection{Plane-Plane Measurement Jacobians}
	\label{pltplj}
	The Jacobian of the angle measurement residual is 
	\begin{equation}
		\Jbf_{\hat\xbf_k}^{\rbf_{kk'}^\alpha}=
		% \frac{\partial\tilde \alpha_{kk'}}{\partial \tilde\pbf_{k}} = 
		\frac{(\hat{\nbf}_{k'})^\top\left(
			%		{^G}\hat{\mathbf{n}}_{\pi}^{\top G} \hat{\mathbf{n}}_{\pi}
			\mathbf{I}_{3}-^{G} \hat{\mathbf{n}}_{k}^{G} \hat{\mathbf{n}}_{k}^{\top}\right)}{^G\hat{d}_{k}}.
	\end{equation}
	The distance measurement residual is calculated as 
	\begin{equation}
		\Jbf_{\hat\xbf_k}^{\rbf_{kk'}^d}=
		\frac{{^G}\hat{\pbf}_{k} -{^G}\hat{\pbf}_{k'}}{\|{^G}\hat{\pbf}_{k} -{^G}\hat{\pbf}_{k'}\|}
		% \frac{\partial \tilde d_{kk'}}{\partial \tilde \pbf_k} = 
		% \frac{\hat{\dbf}_{kk'}}{\|\hat{\dbf}_{kk'}\|} (-{^G}\hat{\nbf}_k{^G}\hat{\nbf}_k^T).
	\end{equation}
	The Jacobian w.r.t. the plane $k'$ can be calculated similarly.
	
	\subsubsection*{Acknowledgments}
	The work is supported by National Research Foundation (NRF) Singapore, ST Engineering-NTU Corporate Lab under its NRF Corporate Lab@ University Scheme.
	
	\bibliographystyle{apalike}
	\bibliography{my_bib_v2.bib}
	
\end{document}